%% file: main.tex
\documentclass{article} 
\usepackage{iclr2027_conference,times}

\input{math_commands.tex}

\usepackage{amssymb}
\usepackage{hyperref}
\usepackage{url}
\usepackage{booktabs}
\usepackage{multirow}
\usepackage{wrapfig}
\usepackage{graphicx}
\usepackage{capt-of}
\usepackage{xspace}
\title{Sonic-AMG: Streaming Human Motion\\Generation from Spatial Audio}
\input{authors}
\hypersetup{
  pdftitle={Sonic-AMG: Streaming Human Motion Generation from Spatial Audio},
  pdfauthor={Jingcheng Ni, Zekun Li, Zhiyang Dou, Xiaoyan Cong, Ziyao Zeng, Hongyang Du, Randall Balestriero, Taku Komura, Srinath Sridhar},
  hidelinks
}

\iclrfinalcopy
\newcommand{\name}{Sonic-AMG\xspace}

\begin{document}
\maketitle
\fancyhead{}
\renewcommand{\headrulewidth}{0pt}

\input{figures/teaser}

\begin{abstract}
\input{sections/abstract}
\end{abstract}

\input{sections/introduction}

\input{sections/related_work}

\input{sections/method}

\begingroup
\setlength{\floatsep}{8pt plus 2pt minus 2pt}
\setlength{\textfloatsep}{12pt plus 2pt minus 2pt}
\input{sections/experiments}

\input{sections/conclusion}
\par
\endgroup

\clearpage
\subsection*{AI use statement}
Generative AI tools were used to polish the manuscript and assist with code development.
The authors take full responsibility for the final manuscript and code.

\subsection*{Reproducibility statement}
Section~\ref{sec:method} describes the model and training objectives, while
Section~\ref{sec:experiments} specifies the evaluation protocols and metrics.
Appendices~\ref{sec:implementation} and~\ref{sec:training_sequences} provide
implementation and training details. Runtime measurement procedures are
described in Appendix~\ref{sec:streaming_efficiency}.
Our code and model will be released upon acceptance.

\bibliography{iclr2027_conference}
\bibliographystyle{iclr2027_conference}

\clearpage
\appendix
\input{sections/appendix}

\end{document}

%% file: math_commands.tex
\usepackage{amsmath,amsfonts,bm}

\def\eqref#1{equation~\ref{#1}}

\def\1{\bm{1}}

\DeclareMathAlphabet{\mathsfit}{\encodingdefault}{\sfdefault}{m}{sl}
\SetMathAlphabet{\mathsfit}{bold}{\encodingdefault}{\sfdefault}{bx}{n}



%% file: authors.tex
\usepackage{latexml}
\iflatexml
\author{%
  Jingcheng Ni\\Brown University
  \and Zekun Li\\Brown University
  \and Zhiyang Dou\\MIT
  \and Xiaoyan Cong\\Brown University
  \and Ziyao Zeng\\Yale University
  \and Hongyang Du\\Brown University
  \and Randall Balestriero\\Brown University
  \and Taku Komura\\The University of Hong Kong
  \and Srinath Sridhar\\Brown University
}
\else
\author{%
  \textbf{Jingcheng Ni}$^{1}$ \quad \textbf{Zekun Li}$^{1}$ \quad \textbf{Zhiyang Dou}$^{2}$ \quad \textbf{Xiaoyan Cong}$^{1}$ \quad \textbf{Ziyao Zeng}$^{3}$\\
  \textbf{Hongyang Du}$^{1}$ \quad \textbf{Randall Balestriero}$^{1}$ \quad \textbf{Taku Komura}$^{4}$ \quad \textbf{Srinath Sridhar}$^{1}$\thanks{Corresponding author.}\\
  $^{1}$ Brown University\\
  $^{2}$ MIT\\
  $^{3}$ Yale University\\
  $^{4}$ The University of Hong Kong
}
\fi

%% file: figures/teaser.tex
\begin{minipage}{\textwidth}
  \centering
  \vspace{-1cm}
  \includegraphics[width=\linewidth]{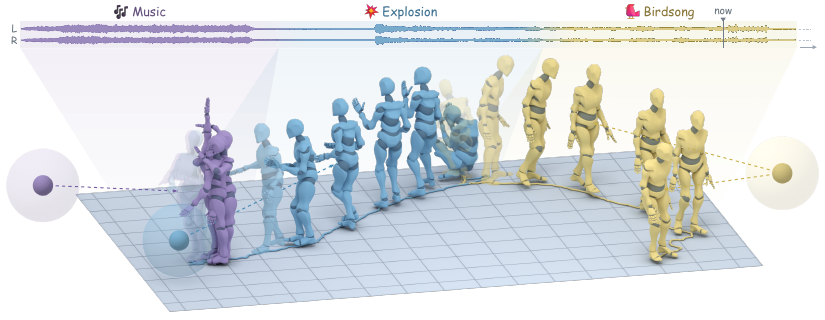}
  \vspace{-1cm}
  \captionof{figure}{\textbf{Streaming motion generation from spatial audio.}
  \name generates continuous reactions to successive sounds, updating each
  source's position relative to the generated body. Colors link sound sources
  and audio segments to the corresponding motions.
  The unified scene shows successive motion snapshots from purple to blue to yellow.
  Translucent bands connect audio segments to their motion responses, while faded
  poses highlight motion transitions between audio segments.}
  \label{fig:teaser}
\end{minipage}
\par\medskip

%% file: sections/abstract.tex
Modeling human reactions to spatial audio requires understanding both what is heard and where it comes from.
Prior work generates motion offline from complete acoustic and spatial input sequences.
Humans, however, react to sounds without knowing what they will hear next.
We present Sonic-AMG, to our knowledge the first framework for streaming human motion generation from spatial audio.
The model encodes acoustic content and body-relative source geometry as separate tokens.
Conditioned on these tokens, a causal transformer with a flow head generates continuous motion latents, from which a causal decoder reconstructs motion segments.
Standard teacher-forced training derives motion histories and body-relative source locations from ground-truth motion, whereas streaming inference relies on generated motion for both, resulting in a training--inference mismatch. To reduce this history discrepancy, we introduce \emph{rehearsal training}, which combines full-sequence teacher-forced supervision with supervised prediction along short autoregressive self-rollouts.
To keep spatial conditioning consistent with generated motion, we propose \emph{spatial feedback}, which recomputes body-relative source locations from generated root poses during both rehearsal training and inference.
On SAM, Sonic-AMG achieves state-of-the-art FID and R-precision.
It further supports real-time online generation of coherent reactions across sound events and source locations.
Project page: \url{https://jingchengni.com/sonic-amg/}.

%% file: sections/introduction.tex
\section{Introduction}
\label{sec:introduction}
Sound provides information about events beyond the visual field, making it an important cue for embodied interaction~\citep{chen2020soundspaces,chen2021semantic}.
Human reactions to sound depend on both its content and the source's location relative to the body.
For example, hearing a voice from behind may prompt a listener to turn toward the speaker before the utterance ends and adjust the movement as new acoustic information arrives.
As the listener turns, the source's direction relative to the body changes. Generating such reactions therefore requires responding as audio arrives and updating spatial conditions as the body moves.

Recent work ~\citep{xu2025mospa} synthesizes complete motion sequences within predefined temporal windows, conditioned on fully observed acoustic and spatial inputs.
Requiring the full input window prevents it from responding causally to an incoming audio stream.
Autoregressive motion models provide a basis for streaming generation by predicting successive motion segments from preceding movements~\citep{shi2024amdm,xiao2025motionstreamer}.
Extending autoregressive learning to spatial audio requires representations that capture both acoustic content and spatial cues.
Furthermore, streaming generation requires each motion prediction to incorporate the acoustic and spatial information currently available as the input evolves.

\begin{wraptable}{r}{0.56\textwidth}
 \centering
 \caption{\textbf{Method comparison.} Online SSL denotes spatial conditioning updated from generated motion.}
 \label{tab:intro_capabilities}
 \small
 \input{tables/intro_capabilities}
\end{wraptable}
We present \name, to our knowledge the first framework for streaming human motion generation from continuous spatial audio (Table~\ref{tab:intro_capabilities}).
Motion is represented in the continuous latent space of a pretrained causal temporal VAE~\citep{xiao2025motionstreamer}. Given a binaural audio stream and the world-space sound source location (SSL), the model
encodes acoustic content and body-relative source geometry as separate tokens.
At each step, a causal transformer processes the incoming acoustic and spatial conditions together with preceding motion latents.
Its output conditions a flow-matching head that samples the next motion latent, and a causal decoder reconstructs the corresponding motion segment.
Repeating this process extends the motion as audio arrives, without requiring the full audio in advance or revising previously generated frames.

We treat acoustic content and spatial cues as distinct modalities of spatial-audio conditioning.
We hypothesize that early fusion introduces interference between them.
\name therefore encodes them as dedicated \emph{what} and \emph{where} tokens, respectively, and learns their interactions within the causal transformer.
With sound source geometry supplied separately, the acoustic branch can focus on sound content rather than a comprehensive description of the binaural signal.
For instance, simple log-mel features in our model yield better results than MOSPA's 2,272-dimensional audio features~\citep{xu2025mospa}.
To design the \emph{where} token, we examine perceptual research to understand how body-relative SSL influences human reactions.
We develop an 11-dimensional spatial event representation inspired by studies of nearby auditory space~\citep{graziano1999nearby}, sound-source motion~\citep{warren2002soundmotion}, and looming~\citep{neuhoff1998rising}.

With acoustic and spatial representations in place, learning the causal generator further requires addressing two sources of \textit{training--inference mismatch}.
The first concerns motion history: teacher-forced training uses ground-truth histories, whereas autoregressive inference relies on the model's own predictions.
To reduce this mismatch, two-forward training replaces selected ground-truth inputs with predictions from a preliminary teacher-forced pass~\citep{mihaylova2019scheduled,xiao2025motionstreamer}.
However, these replacements are still predicted under ground-truth histories, where later replacements do not depend on earlier sampled outputs.
We introduce \emph{rehearsal training}, which combines teacher-forced supervision with supervised prediction on short histories generated autoregressively by the model.
Training on self-rollouts improves robustness to accumulated prediction errors, while teacher-forced supervision provides a stable learning signal across the full sequence.

The second mismatch arises in spatial conditioning. While acoustic content is directly supplied by the incoming audio stream, body-relative SSL depends on the root position and orientation.
MOSPA uses complete SSL sequences precomputed from recorded motion, which encode information about future ground-truth root motion.
Such conditioning is incompatible with causal streaming generation, which requires SSL to be updated from the generated root.
Training with SSL derived from ground-truth motion therefore creates a mismatch with the spatial conditions encountered at inference.
Our experiments show that this mismatch substantially degrades generation quality.
We therefore introduce \emph{spatial feedback}, which recomputes body-relative source coordinates after each generated segment during both rehearsal training and inference.
The updated coordinates condition the next prediction, keeping spatial conditions consistent with the generated motion history.

On SAM~\citep{xu2025mospa}, \name achieves state-of-the-art FID and R-precision.
It further supports causal and online motion generation from streaming spatial audio in real time.
Although trained on individual audio--motion clips, \name can generate coherent motion across successive sound events with varying content and source locations (Fig.~\ref{fig:teaser}), demonstrating robust streaming generation.
Our contributions are:

\noindent\textbullet\enspace We present \name, to our knowledge the first framework for causal human motion generation from streaming spatial audio, with dedicated \emph{what} and \emph{where} tokens for acoustic content and \mbox{spatial cues}.

\noindent\textbullet\enspace We introduce \emph{rehearsal training} to address the history mismatch between training and inference, combining teacher-forced supervision with supervised prediction from histories constructed through short autoregressive self-rollouts.

\noindent\textbullet\enspace We propose \emph{spatial feedback} to address the mismatch in spatial conditioning, updating body-relative source coordinates from generated motion during both rehearsal training and inference.

%% file: tables/intro_capabilities.tex
\setlength{\tabcolsep}{1pt}
\begin{tabular*}{\linewidth}{@{\extracolsep{\fill}}lccc@{}}
\toprule
Method & Spatial Audio & Causal & Online SSL \\
\midrule
EDGE~\citep{tseng2023edge} & $\times$ & $\times$ & --- \\
POPDG~\citep{luo2024popdg} & $\times$ & $\times$ & --- \\
LODGE~\citep{li2024lodge} & $\times$ & $\times$ & --- \\
MOSPA~\citep{xu2025mospa} & $\checkmark$ & $\times$ & $\times$ \\
\textbf{Ours} & $\checkmark$ & $\checkmark$ & $\checkmark$ \\
\bottomrule
\end{tabular*}

%% file: sections/related_work.tex
\section{Related Work}
\label{sec:related_work}

\textbf{Audio-conditioned motion generation.}
Music-driven dance methods use autoregressive models to align motion with musical structure, either in pose space or through learned choreographic codes~\citep{li2021aist,siyao2022bailando}.
Diffusion-based approaches support diverse choreography, motion editing, and hierarchical synthesis of long dance sequences~\citep{alexanderson2023listen,tseng2023edge,luo2024popdg,li2024lodge}.
Diffusion-based visual generation and prediction benefit from complementary conditioning modalities~\citep{zhang2023controlnet,mou2024t2iadapter,zeng2026iris}.
Co-speech gesture models combine acoustic, textual, and speaker or style information to generate gestures that match speech content and rhythm~\citep{yoon2020trimodal,ao2023gesturediffuclip}.
Diffusion models and learned motion priors capture gesture diversity while coordinating body-part movements~\citep{zhu2023diffgesture,ao2023gesturediffuclip,liu2024emage,chen2024diffsheg}.
Streaming systems extend these tasks to incoming music and speech~\citep{saleem2026livegesture,chen2026echoavatar,vu2026rolling,ji2026discoforcing,liu2026gesturefar}.
MOSPA~\citep{xu2025mospa} introduces the SAM dataset of human reactions to spatial audio and a diffusion model conditioned on binaural audio, source coordinates, and reaction style.
Its full-sequence conditioning requires future observations, precluding causal reactions to an incoming audio stream.
MoSAT~\citep{xu2026mosat} conditions a latent flow-matching model on text, acoustic features, and an initial body-relative sound-source direction.
Streaming generation remains future work.
Sonic-AMG enables real-time causal motion generation from incoming spatial audio, with spatial feedback updating body-relative source geometry.

\textbf{Causal motion generation.}
Autoregressive (AR) models use next-token prediction for language modeling~\citep{brown2020language} and image synthesis with discrete visual tokens~\citep{esser2021taming}.
Continuous latent representations support diffusion-based and flow-based generation~\citep{li2024mar,ren2025flowar,zheng2026rae,du2026fusereg}.
Autoregressive approaches pair a transformer with a conditional diffusion or flow head, including in multimodal language models~\citep{sun2026latentlm}.
In motion synthesis, discrete-token models predict quantized motion codes~\citep{zhang2023t2mgpt,jiang2023motiongpt}, whereas AR diffusion models sample successive frames or motion primitives~\citep{shi2024amdm,zhao2025dartcontrol,tevet2025closd,zhang2025primal,zhao2026ardy}.
Streaming generation in latent space requires decoding segments without future latents.
Systems combining causal decoding with transformer-conditioned generative heads support online text control, two-character interaction, and co-speech gestures~\citep{xiao2025motionstreamer,cen2025readytoreact,li2026llamo,liu2026gesturefar}.
Causal decoding alone does not resolve exposure bias: teacher forcing uses recorded histories, whereas inference relies on model predictions.
Diffusion forcing exposes models to contexts at varying noise levels~\citep{chen2024diffusionforcing,cai2026flooddiffusion,ji2026discoforcing}, while other strategies introduce model-generated histories~\citep{bengio2015scheduled,zhao2025dartcontrol,huang2025selfforcing}.
Two-forward training preserves parallel computation using replacement inputs from a preliminary teacher-forced pass~\citep{mihaylova2019scheduled,xiao2025motionstreamer}.
These replacements do not recursively depend on earlier sampled outputs.
Our rehearsal training retains teacher forcing alongside short autoregressive self-rollouts. The combined objective accounts for error propagation across generated segments while retaining sequence-wide supervision.

%% file: sections/method.tex
\section{Method}
\label{sec:method}

\begin{figure}[t]
 \centering
 \includegraphics[width=\linewidth]{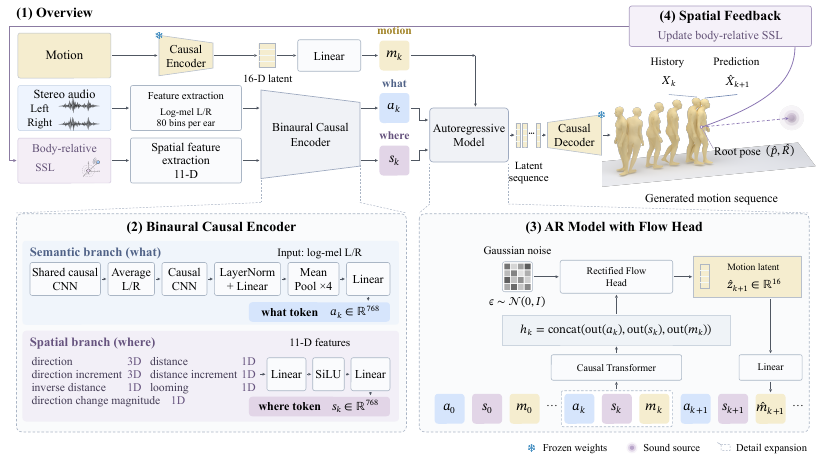}
 \vspace{-0.8cm}
 \caption{\textbf{Sonic-AMG architecture.} A causal transformer and flow head generate motion latents from interleaved acoustic (\emph{what}), spatial (\emph{where}), and motion tokens. Spatial feedback updates body-relative SSL from decoded root poses. $X_k$, $z_k$, and $m_k$ denote a four-frame motion block, its latent, and its projected token, respectively.}
 \label{fig:pipeline}
\end{figure}

Given an incoming binaural audio stream $A$ and a known world-space sound-source position, Sonic-AMG generates motion autoregressively from an initial motion segment, without access to future audio or motion (Fig.~\ref{fig:pipeline}).
We represent motion with continuous latents and encode acoustic content and body-relative source geometry as separate conditioning tokens (Sec.~\ref{sec:representations}).
A causal transformer conditions a flow head to generate the next motion latent from these tokens and the motion history (Sec.~\ref{sec:generator}).
We train the generator using rehearsal training, which combines full-sequence teacher-forced supervision with supervised prediction under histories constructed through short autoregressive self-rollouts (Sec.~\ref{sec:rollout}).
We further apply spatial feedback to update body-relative SSL from the generated root pose during both rehearsal training and inference, keeping spatial conditions consistent with the generated motion history (Sec.~\ref{sec:online_conditions}).

\subsection{Motion and Conditioning Representations}
\label{sec:representations}
\label{sec:conditioning}

\paragraph{Motion representation.}
We represent motion at 30\,Hz.
Each frame $\mathbf{x}_t\in\mathbb{R}^{300}$ comprises the global 3D positions, local rotations in a 6D representation, and 3D velocities of 25 joints~\citep{xu2025mospa}.
We adopt MotionStreamer's causal temporal VAE~\citep{xiao2025motionstreamer} and pretrain it on SAM~\citep{xu2025mospa}.
The VAE encodes motion into a sequence of $K$ latents $\mathbf{z}_k\in\mathbb{R}^{d_z}$ with $d_z=16$ at a temporal stride of $r=4$.
Each latent $\mathbf{z}_k$ corresponds to a four-frame motion block $\mathcal{I}_k=\{rk,\ldots,rk+r-1\}$.
The encoder and decoder remain frozen during generator training.

For spatial audio, we regard acoustic content and source geometry as two distinct conditioning modalities.
We hypothesize that the two modalities can act as noise for one another when fused into a single conditioning representation.
We therefore encode them as separate tokens.

\paragraph{Semantic branch: what.}
The semantic branch in Figure~\ref{fig:pipeline}(2) learns acoustic features for motion prediction from the incoming binaural stream.
We represent each ear's spectral content with 80-dimensional log-mel features and process them with a shared causal temporal CNN.
The shared encoder maps both ears into a common feature space, where we average the two streams to emphasize sound content shared across ears.
A two-layer causal temporal CNN then aggregates recent acoustic context, followed by layer normalization and a linear projection.
Mean pooling over each four-frame block produces $\mathbf{f}_k$ at the motion latent rate, so that each acoustic token aligns with a motion block.
We obtain the acoustic token as $\mathbf{a}_k=W_a\mathbf{f}_k+\mathbf{e}_c$, where $W_a$ is a learned linear projection and $\mathbf{e}_c$ is a learned embedding indexed by SAM's annotated reaction-style label $c$.

\paragraph{Spatial branch: where.}
The spatial branch in Figure~\ref{fig:pipeline}(2) encodes source location and relative motion.
Our descriptor reflects perceptually relevant aspects of spatial sound: proximity~\citep{graziano1999nearby}, relative motion~\citep{warren2002soundmotion}, and approach~\citep{neuhoff1998rising}.
We assume a sound source at known world-space position $\mathbf{g}$.
For root position $\mathbf{p}_t$ and orientation $\mathbf{R}_t\in SO(3)$, the egocentric source location is $\boldsymbol{\ell}_t=\mathbf{R}_t^\top(\mathbf{g}-\mathbf{p}_t)$.
We average source locations within each block, $\overline{\boldsymbol{\ell}}_k=r^{-1}\sum_{t\in\mathcal{I}_k}\boldsymbol{\ell}_t$, and define distance $\rho_k=\|\overline{\boldsymbol{\ell}}_k\|_2$ and direction $\mathbf{d}_k=\overline{\boldsymbol{\ell}}_k/\rho_k$.
The resulting 11-dimensional descriptor contains direction, distance, their increments, and three derived features:
\begin{equation}
 \boldsymbol{\phi}_k=\left[
 \mathbf{d}_k,\rho_k,\Delta\mathbf{d}_k,\Delta\rho_k,
 \rho_k^{-1},\|\Delta\mathbf{d}_k\|_2,
 \frac{\max(-\Delta\rho_k,0)}{\rho_k}\right].
 \label{eq:spatial_descriptor}
\end{equation}
For $k>0$, the direction increment $\Delta\mathbf{d}_k=\mathbf{d}_k-\mathbf{d}_{k-1}$ and distance increment $\Delta\rho_k=\rho_k-\rho_{k-1}$ measure changes between consecutive motion blocks.
The final three features are inverse distance, direction change magnitude, and looming, respectively.
Inverse distance emphasizes proximity, while direction change magnitude summarizes changes in source bearing.
We represent looming by the decrease in distance per block normalized by the current distance, with zero assigned when the distance does not decrease.
The spatial token is $\mathbf{s}_k=E_s((\boldsymbol{\phi}_k-\boldsymbol{\mu}_\phi)/\boldsymbol{\sigma}_\phi)+\mathbf{e}_c$, where $E_s$ is a two-layer MLP and $\boldsymbol{\mu}_\phi,\boldsymbol{\sigma}_\phi$ are the training-set mean and standard deviation.
At inference, acoustic tokens are computed from the incoming audio, while spatial feedback updates spatial tokens from the generated body state (Section~\ref{sec:online_conditions}).

\subsection{Causal Motion Generation}
\label{sec:formulation}
\label{sec:generator}

Given an initial motion latent $\mathbf{z}_0$, we predict each subsequent latent $\mathbf{z}_{k+1}$ from the motion history $\mathbf{z}_{0:k}$, conditioned on acoustic and spatial tokens up to block $k$:
\begin{equation}
 p_\theta(\mathbf{z}_{1:K-1}\mid A,\mathbf{g},c,\mathbf{z}_0)
 =\prod_{k=0}^{K-2}p_\theta(\mathbf{z}_{k+1}\mid
 \mathbf{z}_{0:k},\mathbf{a}_{0:k},\mathbf{s}_{0:k},c),
 \label{eq:causal_factorization}
\end{equation}
where $\theta$ denotes the generator parameters.

We use a learned linear projection $W_m$ to obtain motion tokens $\mathbf{m}_k=W_m\mathbf{z}_k$ and feed the interleaved sequence to a causal transformer (Figure~\ref{fig:pipeline}(3)).
We concatenate the transformer outputs at the three token positions of block $k$ to form the conditioning vector $\mathbf{h}_k$ for the flow head:
\begin{equation}
 \mathbf{U}_{0:k}=[\mathbf{a}_0,\mathbf{s}_0,\mathbf{m}_0,\ldots,
 \mathbf{a}_k,\mathbf{s}_k,\mathbf{m}_k],\qquad
 \mathbf{h}_k=[\mathbf{o}^a_k;\mathbf{o}^s_k;\mathbf{o}^m_k],
 \label{eq:interleaving}
\end{equation}
where $\mathbf{o}^a_k,\mathbf{o}^s_k,\mathbf{o}^m_k$ denote the corresponding transformer outputs.
At inference, we reuse cached keys and values for the history and process only the new token triplet at each step.

Conditioned on $\mathbf{h}_k$, a rectified flow head samples the next motion latent from Gaussian noise.
The head predicts a velocity field $v_\theta$, which we train with conditional flow matching~\citep{liu2023rectified,lipman2023flowmatching}.
For a next-latent target $\mathbf{z}$ and conditioning vector $\mathbf{h}$, the interpolation at flow time $\tau\in[0,1]$ is $\mathbf{u}_\tau=(1-\tau)\boldsymbol{\epsilon}+\tau\mathbf{z}$, where $\boldsymbol{\epsilon}\sim\mathcal{N}(0,I)$.
We sample $\tau$ from a logit-normal distribution and minimize
\begin{equation}
 \ell_{\mathrm{FM}}(\mathbf{z},\mathbf{h})=
 \mathbb{E}_{\boldsymbol{\epsilon},\tau}\left[
 \frac{1}{d_z}\left\|v_\theta(\mathbf{u}_\tau,\tau;\mathbf{h})
 -(\mathbf{z}-\boldsymbol{\epsilon})\right\|_2^2\right].
 \label{eq:flow_loss}
\end{equation}
Sampling integrates the learned velocity field from $\tau=0$ to $1$ with ten Euler steps.
We write the sampled latent as $\widehat{\mathbf{z}}_{k+1}=G_\theta(\boldsymbol{\epsilon};\mathbf{h}_k)$, where $G_\theta$ denotes the flow sampler.
At inference, classifier-free guidance~\citep{ho2022cfg} combines velocity predictions with and without the current acoustic and spatial conditions.

We project the sampled latent $\widehat{\mathbf{z}}_{k+1}$ to a motion token and append it to the motion history.
The causal decoder reconstructs the new motion segment from the extended latent sequence without revising earlier frames.
Spatial feedback then recomputes body-relative source geometry from the decoded root pose for the next prediction (Section~\ref{sec:online_conditions}).

\subsection{Rehearsal training}
\label{sec:rollout}

In two-forward training~\citep{mihaylova2019scheduled,xiao2025motionstreamer}, an initial teacher-forced pass estimates motion latents from GT histories.
Selected estimates then replace GT motion inputs for a second supervised pass.
Although this exposes the model to predicted inputs, those inputs are not generated recursively from earlier predictions, as they would be during autoregressive inference.

Rehearsal training instead constructs motion histories through short autoregressive self-rollouts.
Each sampled latent extends the history used for the next prediction.
We combine next-latent supervision under these recursively generated histories with full-sequence teacher-forced supervision.
Both objectives use the corresponding GT next latent as the prediction target.

\paragraph{Teacher-forced prediction and initialization.}
A teacher-forced pass over the full sequence yields the loss
\begin{equation}
 \mathcal{L}_{\mathrm{TF}}=\frac{1}{K-1}\sum_{k=0}^{K-2}
 \ell_{\mathrm{FM}}(\mathbf{z}^{\mathrm{GT}}_{k+1},\mathbf{h}^{\mathrm{GT}}_k).
 \label{eq:reference_loss}
\end{equation}
Here, $\mathbf{h}^{\mathrm{GT}}_k$ is computed from the GT motion history $\mathbf{z}^{\mathrm{GT}}_{0:k}$ and the corresponding acoustic and spatial tokens.
We sample a start $s$ for an $H$-transition rollout independently for each sequence.
A detached denoising estimate from the same teacher-forced pass initializes $\overline{\mathbf{z}}_s$.
The transformer cache and acoustic context are reset at $s$.

\paragraph{Recursive context construction.}
At step $k$, the rollout history $\overline{\mathbf{z}}_{s:k}$ and the corresponding acoustic and spatial token sequences determine $\overline{\mathbf{h}}_k$.
We compute the flow-matching loss against $\mathbf{z}^{\mathrm{GT}}_{k+1}$ under this context.
We then independently sample a latent under the same context and append it to the rollout history:
\begin{equation}
 \overline{\mathbf{z}}_{k+1}
 =\operatorname{sg}\!\left[G_\theta(\boldsymbol{\epsilon}_k;\overline{\mathbf{h}}_k)\right],
 \qquad
 \overline{\mathbf{z}}_{s:k+1}
 =[\overline{\mathbf{z}}_{s:k},\overline{\mathbf{z}}_{k+1}],
 \label{eq:rehearsal_recursion}
\end{equation}
where $\operatorname{sg}$ stops gradients and $\boldsymbol{\epsilon}_k$ is fresh Gaussian noise.
We average the flow-matching losses over the $H$ transitions and combine the two objectives with equal weight:
\begin{equation}
 \mathcal{L}_{\mathrm{reh}}=\frac{1}{H}\sum_{k=s}^{s+H-1}
 \ell_{\mathrm{FM}}(\mathbf{z}^{\mathrm{GT}}_{k+1},\overline{\mathbf{h}}_k),
 \qquad
 \mathcal{L}=\tfrac12(\mathcal{L}_{\mathrm{TF}}+\mathcal{L}_{\mathrm{reh}}).
 \label{eq:combined_loss}
\end{equation}
The teacher-forced term supervises the full recorded sequence, while the rollout term supervises continuations of increasingly long sampled histories.
We use $H=16$ by default and unguided sampling during rehearsal.
Gradients from both objectives update the conditioning encoders, transformer, and flow head.
Gradients propagate through cached transformer states but not through sampled latents or the sampling procedure.

\subsection{Spatial feedback}
\label{sec:online_conditions}

Spatial feedback recomputes the source's body-relative coordinates from generated motion during both training rollouts and inference (Figure~\ref{fig:pipeline}(4)).
Given the decoder input prefix $\mathbf{z}_{0:k}$, the causal decoder $\mathcal{D}_\psi$ with frozen parameters $\psi$ reconstructs block $k$, whose root positions $\widehat{\mathbf{p}}_t$ and orientations $\widehat{\mathbf{R}}_t$ determine the updated source coordinates:
\begin{equation}
 \widehat{\mathbf{x}}_{\mathcal{I}_k}
 =\left.\mathcal{D}_\psi(\mathbf{z}_{0:k})\right|_{\mathcal{I}_k},
 \qquad
 \widehat{\boldsymbol{\ell}}_t
 =\widehat{\mathbf{R}}_t^\top(\mathbf{g}-\widehat{\mathbf{p}}_t),
 \quad t\in\mathcal{I}_k.
 \label{eq:spatial_update}
\end{equation}
We average $\widehat{\boldsymbol{\ell}}_t$ within block $k$ and recompute the descriptor in Eq.~(\ref{eq:spatial_descriptor}) from the current and preceding pooled locations.
During training rollouts, the recomputed geometry is detached before spatial encoding.
The spatial encoder in Section~\ref{sec:representations} maps this descriptor to $\mathbf{s}_k$, which conditions the prediction of block $k+1$.

%% file: sections/experiments.tex
\section{Experiments}
\label{sec:experiments}


\textbf{Dataset.}
We evaluate on SAM~\citep{xu2025mospa}, which pairs spatial audio with human reactions. Autoregressive models receive four initial ground-truth motion frames and the reaction style.

\textbf{Spatial conditioning inputs.}
We consider two spatial-conditioning protocols: \textbf{GT} uses body-relative SSL precomputed from recorded motion, whereas \textbf{Online} recomputes SSL from generated root poses during inference. GT SSL relies on ground-truth root information that is unavailable during causal streaming generation. We include a GT SSL variant to enable fair comparison with MOSPA and other baselines under their established evaluation protocol. Online evaluation is more challenging, as inaccuracies in generated root poses perturb the SSL used to condition subsequent predictions, introducing an additional pathway for error propagation. All qualitative results of Sonic-AMG shown in the experiments are generated under the Online protocol. 

Following the established SAM evaluation protocol, SSL inputs are available to all methods in our main comparisons. Such inputs can be supplied by dataset annotations, an upstream localization module~\citep{krause2024binaural,krause2024distance}, or virtual-scene metadata. Our modular conditioning design also supports a No-SSL variant that derives spatial tokens directly from binaural audio, without requiring SSL inputs at inference, detailed in Appendix~\ref{sec:modular_spatial_conditioning}.

\textbf{Evaluation metrics.}
Following MOSPA~\citep{xu2025mospa}, we report Fr\'echet Inception Distance (FID), R-precision (R@1--3), and feature-space diversity. Lower FID indicates better distributional fidelity, while higher R-precision indicates better conditioning alignment. Diversity closer to real motion is preferred. We report 95\% confidence intervals. Bold and underlined values mark the best and second-best results within each comparison.

\subsection{Comparison with Existing Methods}
\label{sec:reference_results}

\begin{table}[tbp]
 \centering
 \caption{\textbf{Quantitative comparison on SAM.} Causal denotes support for causal generation. $\dagger$ denotes results from \citet{xu2025mospa}.}
 \label{tab:reference_results}
 \small
 \input{tables/main_results}
\end{table}

\textbf{Quantitative comparison.}
Sonic-AMG supports real-time streaming generation (Section~\ref{sec:streaming_results}) while achieving the best distributional fidelity and conditioning alignment among the compared methods under the GT SSL protocol (Table~\ref{tab:reference_results}). Under the Online protocol, Sonic-AMG maintains strong distributional fidelity and conditioning alignment while updating spatial conditions from its own generated motion. We attribute these results to the complementary roles of our separate conditioning representations and rehearsal training. Dedicated conditioning tokens allow each motion prediction to distinguish acoustic content from body-relative source geometry, supporting reactions that account for both cues. Rehearsal training combines sequence-wide supervision with local autoregressive consistency. Teacher forcing provides supervision across complete motion sequences, supporting motion fidelity and long-range temporal structure, while short self-rollouts expose the model to the temporal consequences of its own predictions. The joint objective directly targets the recursive prediction process required for streaming generation, which is absent from MOSPA's full-sequence training. The ablations in Sections~\ref{sec:feedback_results} and~\ref{sec:representation_results} support the contributions of the representation and training.

\begin{figure}[!t]
 \centering
 \includegraphics[width=0.95\linewidth]{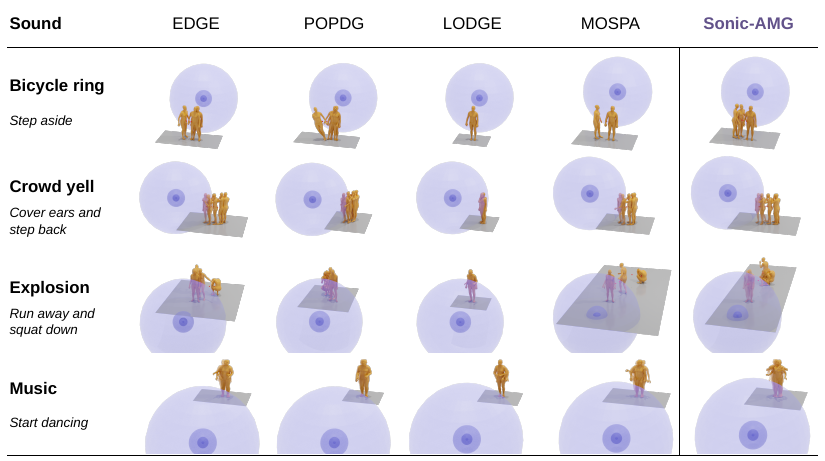}
 \vspace{-0.3cm}
 \caption{\textbf{Qualitative comparison on SAM.} EDGE, POPDG, and LODGE panels are reproduced from \citet{xu2025mospa}; purple markers indicate sound sources.
 }
 \label{fig:qualitative_baselines}
\end{figure}

\textbf{Qualitative comparison.}
\label{sec:qualitative}
Sonic-AMG produces recognizable and spatially appropriate reactions under the Online protocol (Fig.~\ref{fig:qualitative_baselines}). It covers the ears in response to crowd noise and combines movement away from the source with a lowered posture for an explosion, whereas several baselines exhibit limited postural change or displacement in these examples. These comparisons illustrate the model's ability to coordinate the type of reaction with movement relative to the sound source. Additional visual results are provided in Appendix~\ref{sec:additional_visual_results}, with videos available in the supplementary material.

\begin{figure}[!t]
 \centering
 \input{figures/qualitative_grid}
 \vspace{-0.3cm}
 \caption{\textbf{Reactions over the full audio duration.} MOSPA (top) generates 8\,s of motion; Sonic-AMG (bottom) covers the full 10--12\,s audio clips. Six poses are sampled uniformly over each sequence's own duration. Purple markers denote sound sources; red dots trace root trajectories.}
 \label{fig:qualitative}
\end{figure}
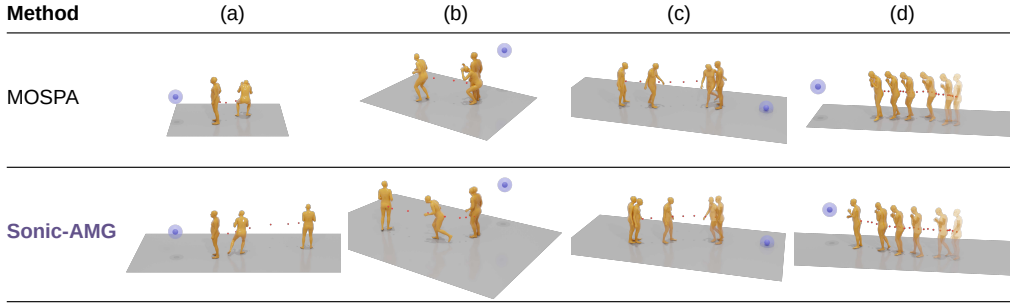

\textbf{Generation over complete audio clips.}
Sonic-AMG generates complete reactions to audio clips of different durations using a single trained model (Fig.~\ref{fig:qualitative}). While MOSPA is restricted to a fixed output window, Sonic-AMG extends motion as audio arrives. Postural changes and body displacement can therefore unfold throughout the audio, yielding coherent and complete reaction sequences.

\textbf{Physical simulation.}
We retarget reactions to Unitree G1 with GMR~\citep{araujo2026gmr} and track them with pretrained GEAR-SONIC~\citep{luo2026sonic} in MuJoCo~\citep{todorov2012mujoco} (Fig.~\ref{fig:physical_simulation}).

\begin{figure}[!t]
 \centering
 \input{figures/physical_simulation}
 \vspace{-0.3cm}
 \caption{\textbf{Physical simulation of generated reactions.} See Appendix~\ref{sec:physics_simulation} for details.}
 \label{fig:physical_simulation}
\end{figure}

\subsection{Streaming Efficiency and Latency}
\label{sec:streaming_results}

\begin{table}[tbp]
 \centering
 \caption{\textbf{Streaming computation.} All measurements use batch size one. For Sonic-AMG, step times (ms) are measured from precomputed acoustic tokens. Online includes motion decoding and spatial feedback; GT measures latent generation with precomputed SSL. RTF is computation time divided by generated motion duration. Real-time denotes causal streaming with $\mathrm{RTF}<1$. MOSPA has no four-frame step time.}
 \label{tab:streaming_results}
 \small
 \input{tables/streaming_results}
\end{table}

\textbf{Real-time computation.}
Sonic-AMG sustains real-time Online generation on both RTX A6000 and TITAN RTX GPUs (Table~\ref{tab:streaming_results}). Each autoregressive step produces four motion frames at 30\,Hz, allowing 133.3\,ms of computation per block. The 95th-percentile step time remains below this budget on both GPUs, including motion decoding and spatial feedback. MOSPA achieves an RTF below one on RTX A6000, but requires the complete audio window before generation and therefore does not support causal streaming. On TITAN RTX, its RTF exceeds one, while Sonic-AMG retains real-time throughput.

\textbf{Initial response latency.}
On RTX A6000, Sonic-AMG reduces estimated initial response latency from MOSPA's 10{,}857\,ms to approximately 193\,ms, a reduction by a factor of about 56. Causal blockwise generation allows reactions to begin as audio arrives, whereas MOSPA must first collect its complete 8\,s input window. These estimates combine input waiting with mean steady-state generation time. Appendix~\ref{sec:streaming_efficiency} provides the timing protocols, excluded operations, and latency breakdown.

\subsection{Rehearsal training}
\label{sec:feedback_results}

\begin{table}[!t]
 \centering
 \caption{\textbf{Training strategies.} Each group uses the same SSL protocol for predicted training histories and evaluation. Two-forward schedules increase ($0\!\to\!1$), decrease ($1\!\to\!0$), or fix ($p=1$) the fraction of replaced motion inputs.}
 \label{tab:training_results}
 \small
 \input{tables/training_results}
 \vspace{-0.4cm}
\end{table}

\textbf{Comparison with two-forward training.}
Rehearsal training improves motion fidelity and conditioning alignment over all two-forward schedules under both SSL protocols (Table~\ref{tab:training_results}). All variants share the generator architecture, and Online two-forward baselines also update spatial conditions. The gains support recursive history construction, which exposes the model to error propagation across generated segments that teacher-forced replacements do not capture.

\textbf{Role of teacher-forced supervision.}
Self-rollouts are most effective when combined with full-sequence teacher-forced supervision. Removing this supervision degrades fidelity and alignment. Restoring the motion and conditioning prefix in Rollout Only with History (Appendix~\ref{sec:training_sequences}) improves performance but does not recover rehearsal's fidelity or top-rank retrieval under either protocol. Additional context therefore cannot fully replace supervision over complete sequences.

\subsection{Effect of spatial feedback}
\label{sec:condition_gap}
\begin{table}[!t]
 \centering
 \begin{minipage}[t]{0.46\linewidth}
  \centering
 \caption{\textbf{Cross-condition evaluation.}}
 \label{tab:cross_conditions}
 \end{minipage}\hfill
 \begin{minipage}[t]{0.52\linewidth}
  \centering
 \caption{\textbf{Conditioning representations.}}
 \label{tab:spatial_representation}
 \label{tab:acoustic_representation}
 \end{minipage}
 \par
 \begin{minipage}[t]{0.46\linewidth}
  \centering\small
  \input{tables/cross_results}
 \end{minipage}\hfill
 \begin{minipage}[t]{0.52\linewidth}
  \centering\small
  \newsavebox{\crossconditionheightbox}
  \sbox{\crossconditionheightbox}{\input{tables/cross_results}}
  \resizebox*{\linewidth}{\dimexpr\ht\crossconditionheightbox+\dp\crossconditionheightbox\relax}{%
   \input{tables/representation_combined}  }
 \end{minipage}
 \vspace{-0.4cm}
\end{table}

Adapting to generated motion histories alone is insufficient for Online generation. Table~\ref{tab:cross_conditions} compares two rehearsal-trained models. For the GT-trained model, switching evaluation from GT to Online SSL with fixed generation settings substantially degrades motion fidelity and conditioning alignment. During training, it encounters prediction errors in its motion history but not the resulting changes in spatial conditions. Training with spatial feedback substantially improves fidelity and alignment under the same Online protocol. The comparison supports updating body-relative SSL during rehearsal, so that motion histories and spatial conditions reflect the same generated trajectory.

\subsection{Acoustic and Spatial Representation}
\label{sec:representation_results}
\vspace{-0.2cm}

\textbf{Separate spatial tokens.}
All representation ablations use Online SSL. \emph{Fused} linearly projects concatenated acoustic and spatial embeddings into a single conditioning token. Separate tokens improve fidelity and alignment (Table~\ref{tab:spatial_representation}). Preserving distinct inputs allows the transformer to attend to acoustic and spatial cues separately.

\textbf{Learning reaction-relevant acoustic features.}
MOSPA aggregates diverse audio descriptors into a 2,272-dimensional feature vector. Our learned log-mel features achieve better fidelity and conditioning alignment (Table~\ref{tab:acoustic_representation}). This improvement is consistent with our hypothesis that redundant or task-irrelevant acoustic variation can interfere with condition modeling. Separating what from where helps the model associate acoustic events with spatially appropriate responses.





%% file: tables/main_results.tex
\setlength{\tabcolsep}{1pt}
\begin{tabular*}{\linewidth}{@{\extracolsep{\fill}}lccrrrrr@{}}
\toprule
Method & Causal & Eval. SSL & FID$\downarrow$ & R@1$\uparrow$ & R@2$\uparrow$ & R@3$\uparrow$ & Diversity$\rightarrow$ \\
\midrule
GT motion$^{\dagger}$ & --- & --- & 0.001 & 1.000\textsuperscript{$\pm$0.000} & 1.000\textsuperscript{$\pm$0.000} & 1.000\textsuperscript{$\pm$0.000} & 23.616\textsuperscript{$\pm$0.188} \\
\midrule
EDGE$^{\dagger}$ & $\times$ & GT & 13.993 & 0.886\textsuperscript{$\pm$0.005} & 0.960\textsuperscript{$\pm$0.003} & 0.977\textsuperscript{$\pm$0.002} & 23.099\textsuperscript{$\pm$0.196} \\
POPDG$^{\dagger}$ & $\times$ & GT & 20.967 & 0.762\textsuperscript{$\pm$0.006} & 0.886\textsuperscript{$\pm$0.005} & 0.934\textsuperscript{$\pm$0.003} & 22.536\textsuperscript{$\pm$0.170} \\
LODGE$^{\dagger}$ & $\times$ & GT & 102.289 & 0.444\textsuperscript{$\pm$0.006} & 0.594\textsuperscript{$\pm$0.005} & 0.679\textsuperscript{$\pm$0.004} & 21.101\textsuperscript{$\pm$0.141} \\
Bailando$^{\dagger}$ & $\times$ & GT & 168.396 & 0.077\textsuperscript{$\pm$0.003} & 0.134\textsuperscript{$\pm$0.003} & 0.182\textsuperscript{$\pm$0.004} & 17.347\textsuperscript{$\pm$0.247} \\
MOSPA & $\times$ & GT & 8.242 & 0.942\textsuperscript{$\pm$0.004} & \underline{0.984}\textsuperscript{$\pm$0.002} & \underline{0.994}\textsuperscript{$\pm$0.001} & \textbf{23.412}\textsuperscript{$\pm$\textbf{0.169}} \\
Sonic-AMG (Ours) & $\checkmark$ & GT & \textbf{6.505} & \textbf{0.957}\textsuperscript{$\pm$\textbf{0.004}} & \textbf{0.991}\textsuperscript{$\pm$\textbf{0.002}} & \textbf{0.998}\textsuperscript{$\pm$\textbf{0.001}} & \underline{23.241}\textsuperscript{$\pm$0.158} \\

\midrule
Sonic-AMG (Ours) & $\checkmark$ & Online & \underline{8.066} & \underline{0.943}\textsuperscript{$\pm$0.004} & \underline{0.984}\textsuperscript{$\pm$0.002} & \underline{0.994}\textsuperscript{$\pm$0.001} & 23.194\textsuperscript{$\pm$0.149} \\

\bottomrule
\end{tabular*}

%% file: figures/qualitative_grid.tex
\includegraphics[width=0.95\linewidth]{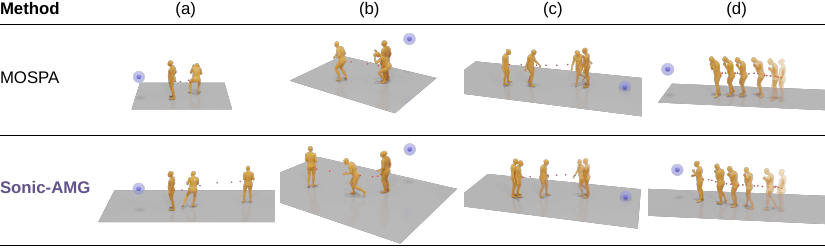}

%% file: figures/physical_simulation.tex
\includegraphics[width=\linewidth]{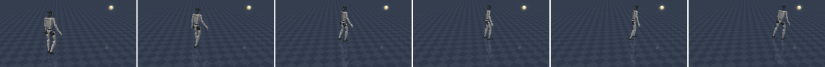}\par
\smallskip
{\small (a) Birdsong: approaching the sound source.}\par
\medskip
\includegraphics[width=\linewidth]{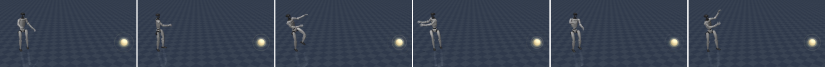}\par
\smallskip
{\small (b) Singing: performing arm and leg gestures.}

%% file: tables/streaming_results.tex
\setlength{\tabcolsep}{3pt}
\begin{tabular*}{\linewidth}{@{\extracolsep{\fill}}lllrrrrc@{}}
\toprule
GPU & Method & Eval. SSL & Mean$\downarrow$ & Median$\downarrow$ & P95$\downarrow$ & RTF$\downarrow$ & Real-time \\
\midrule
\multirow{3}{*}{RTX A6000} & MOSPA & GT & --- & --- & --- & 0.357 & $\times$ \\
 & Ours & GT & 41.2 & 41.1 & 42.5 & 0.309 & $\checkmark$ \\
 & Ours & Online & 50.2 & 50.9 & 53.2 & 0.376 & $\checkmark$ \\
\midrule
\multirow{3}{*}{TITAN RTX} & MOSPA & GT & --- & --- & --- & 1.427 & $\times$ \\
 & Ours & GT & 102.5 & 105.8 & 108.1 & 0.769 & $\checkmark$ \\
 & Ours & Online & 118.0 & 117.9 & 125.6 & 0.885 & $\checkmark$ \\
\bottomrule
\end{tabular*}

%% file: tables/training_results.tex
\setlength{\tabcolsep}{1pt}
\begin{tabular*}{\linewidth}{@{\extracolsep{\fill}}llrrrrr@{}}
\toprule
Training strategy & Eval. SSL & FID$\downarrow$ & R@1$\uparrow$ & R@2$\uparrow$ & R@3$\uparrow$ & Diversity$\rightarrow$ \\
\midrule
\multicolumn{2}{l}{GT motion$^{\dagger}$} & 0.001 & 1.000\textsuperscript{$\pm$0.000} & 1.000\textsuperscript{$\pm$0.000} & 1.000\textsuperscript{$\pm$0.000} & 23.616\textsuperscript{$\pm$0.188} \\
\midrule
Two-forward ($0\!\to\!1$) & GT & 10.797 & 0.839\textsuperscript{$\pm$0.006} & 0.924\textsuperscript{$\pm$0.004} & 0.956\textsuperscript{$\pm$0.003} & 22.856\textsuperscript{$\pm$0.160} \\
Two-forward ($1\!\to\!0$) & GT & \underline{8.001} & 0.926\textsuperscript{$\pm$0.004} & 0.972\textsuperscript{$\pm$0.002} & 0.983\textsuperscript{$\pm$0.001} & 23.106\textsuperscript{$\pm$0.136} \\
Two-forward ($p=1$) & GT & 8.033 & \underline{0.934}\textsuperscript{$\pm$0.004} & \underline{0.981}\textsuperscript{$\pm$0.002} & \underline{0.992}\textsuperscript{$\pm$0.001} & \underline{23.223}\textsuperscript{$\pm$0.172} \\
Rollout Only & GT & 14.854 & 0.860\textsuperscript{$\pm$0.005} & 0.937\textsuperscript{$\pm$0.004} & 0.966\textsuperscript{$\pm$0.003} & 22.550\textsuperscript{$\pm$0.161} \\
Rollout Only with History & GT & 8.585 & 0.915\textsuperscript{$\pm$0.004} & 0.968\textsuperscript{$\pm$0.003} & 0.984\textsuperscript{$\pm$0.003} & 22.883\textsuperscript{$\pm$0.142} \\
Rehearsal (Ours) & GT & \textbf{6.505} & \textbf{0.957}\textsuperscript{$\pm$\textbf{0.004}} & \textbf{0.991}\textsuperscript{$\pm$\textbf{0.002}} & \textbf{0.998}\textsuperscript{$\pm$\textbf{0.001}} & \textbf{23.241}\textsuperscript{$\pm$\textbf{0.158}} \\
\midrule
Two-forward ($0\!\to\!1$) & Online & 15.367 & 0.796\textsuperscript{$\pm$0.005} & 0.887\textsuperscript{$\pm$0.004} & 0.923\textsuperscript{$\pm$0.003} & 22.974\textsuperscript{$\pm$0.147} \\
Two-forward ($1\!\to\!0$) & Online & 12.695 & 0.846\textsuperscript{$\pm$0.005} & 0.933\textsuperscript{$\pm$0.003} & 0.963\textsuperscript{$\pm$0.002} & \underline{23.239}\textsuperscript{$\pm$0.151} \\
Two-forward ($p=1$) & Online & \underline{11.766} & 0.887\textsuperscript{$\pm$0.004} & 0.958\textsuperscript{$\pm$0.003} & \textbf{0.981}\textsuperscript{$\pm$\textbf{0.002}} & 23.195\textsuperscript{$\pm$0.166} \\
Rollout Only & Online & 17.036 & 0.792\textsuperscript{$\pm$0.006} & 0.903\textsuperscript{$\pm$0.005} & 0.945\textsuperscript{$\pm$0.003} & 22.660\textsuperscript{$\pm$0.151} \\
Rollout Only with History & Online & 12.320 & \underline{0.891}\textsuperscript{$\pm$0.005} & \underline{0.961}\textsuperscript{$\pm$0.002} & \underline{0.980}\textsuperscript{$\pm$0.002} & 22.684\textsuperscript{$\pm$0.110} \\
Rehearsal (Ours) & Online & \textbf{8.714} & \textbf{0.909}\textsuperscript{$\pm$\textbf{0.005}} & \textbf{0.966}\textsuperscript{$\pm$\textbf{0.002}} & \textbf{0.981}\textsuperscript{$\pm$\textbf{0.001}} & \textbf{23.353}\textsuperscript{$\pm$\textbf{0.189}} \\
\bottomrule
\end{tabular*}

%% file: tables/cross_results.tex
\setlength{\tabcolsep}{1pt}
\begin{tabular*}{\linewidth}[t]{@{\extracolsep{\fill}}llrrrrr@{}}
\toprule
\begin{tabular}[c]{@{}l@{}}Train\\SSL\end{tabular} & \begin{tabular}[c]{@{}l@{}}Eval.\\SSL\end{tabular} & FID$\downarrow$ & R@1$\uparrow$ & R@2$\uparrow$ & R@3$\uparrow$ & Div.$\rightarrow$ \\
\midrule
\multicolumn{2}{l}{GT motion$^{\dagger}$} & 0.001 & 1.000 & 1.000 & 1.000 & 23.616 \\
\midrule
GT & GT & 6.505 & 0.957 & 0.991 & 0.998 & 23.241 \\
GT & Online & 25.872 & 0.714 & 0.827 & 0.868 & 22.067 \\
\midrule
\multirow{2}{*}{Online} & \multirow{2}{*}{Online} & \multirow{2}{*}{8.714} & \multirow{2}{*}{0.909} & \multirow{2}{*}{0.966} & \multirow{2}{*}{0.981} & \multirow{2}{*}{23.353} \\
 & & & & & & \\
\bottomrule
\end{tabular*}

%% file: tables/representation_combined.tex
\setlength{\tabcolsep}{0.8pt}
\begin{tabular*}{\linewidth}[t]{@{\extracolsep{\fill}}llrrrrr@{}}
\toprule
\multicolumn{2}{l}{\vphantom{\begin{tabular}[c]{@{}l@{}}Train\\SSL\end{tabular}}Configuration} & FID$\downarrow$ & R@1$\uparrow$ & R@2$\uparrow$ & R@3$\uparrow$ & Div.$\rightarrow$ \\
\midrule
\multicolumn{2}{l}{GT motion$^{\dagger}$} & 0.001 & 1.000 & 1.000 & 1.000 & 23.616 \\
\midrule
\multirow{2}{*}{Spatial} & Fused & \underline{11.445} & \underline{0.881} & \underline{0.951} & \underline{0.972} & \underline{23.083} \\
 & Separate & \textbf{8.714} & \textbf{0.909} & \textbf{0.966} & \textbf{0.981} & \textbf{23.353} \\
\midrule
\multirow{2}{*}{Acoustic} & MOSPA & \underline{12.759} & \underline{0.818} & \underline{0.907} & \underline{0.940} & \underline{22.694} \\
 & Ours & \textbf{8.714} & \textbf{0.909} & \textbf{0.966} & \textbf{0.981} & \textbf{23.353} \\
\bottomrule
\end{tabular*}

%% file: sections/conclusion.tex
\section{Conclusion}
\vspace{-0.2cm}
We presented Sonic-AMG for real-time autoregressive motion generation from streaming spatial audio, with separate acoustic and spatial representations. Rehearsal training combines full-sequence teacher-forced supervision with short autoregressive self-rollouts, while spatial feedback keeps source coordinates consistent with generated motion. Sonic-AMG achieves state-of-the-art FID and R-precision on SAM and generates coherent reactions across sound events and source locations.

%% file: sections/appendix.tex
\section{Limitations and Future Work}
\label{sec:limitations}

Our main experiments assume known world-space source positions during inference. Our No-SSL variant (Appendix~\ref{sec:modular_spatial_conditioning}) provides an initial demonstration that spatial conditioning can be derived from binaural audio. A systematic evaluation of alternative spatial inference methods and their effects on motion generation remains future work. Our evaluation is limited to SAM, so generalization to broader acoustic environments and human reactions remains to be assessed. Evidence for long-horizon coherence and event transitions is primarily qualitative. Future work should establish quantitative benchmarks for evaluating motion quality over longer audio streams with more diverse event transitions.

\input{sections/appendix_visual_results}
\input{sections/appendix_streaming}
\input{sections/appendix_modular_spatial}
\clearpage

\section{Implementation Details}
\label{sec:implementation}

\paragraph{Dataset.}
SAM~\citep{xu2025mospa} contains binaural recordings paired with human reactions across 27 sound scenarios and three reaction styles. 

\paragraph{Architecture and optimization.}
The generator uses a 12-layer, 12-head transformer of width $d=768$, with rotary positional embeddings and query--key normalization. Motion latents have dimension $d_z=16$. The conditional flow head contains nine residual MLP blocks of width 1792 with adaptive layer normalization. We optimize the generator and conditioning modules with AdamW, using batch size 64, learning rate $10^{-5}$, weight decay $10^{-6}$, and momentum coefficients $(0.9,0.95)$. Each supervised transition uses four independently sampled noise--time pairs. Padded prediction positions are excluded from all losses.

\paragraph{Training hardware and time.}
For the $H=16$ configuration, we train the online rehearsal model from scratch for 150{,}000 updates on a single NVIDIA L40S GPU, taking approximately 63 hours. The 130{,}000-update checkpoint is reached after approximately 54 hours. These wall-clock times include periodic validation and checkpointing, but exclude motion-autoencoder pretraining.

\paragraph{Causal motion autoencoder.}
The temporal autoencoder is pretrained using a Gaussian reconstruction objective, KL regularization, and an additional root-position reconstruction term. The encoder and decoder are then fixed during generator training. Motion channels are standardized using training-set statistics; decoded channels are converted back to physical coordinates before source geometry is computed. The initial GT latent encodes four motion frames.

\paragraph{Acoustic encoding and feature availability.}
Audio is sampled at 48\,kHz. Each channel is converted to 80-dimensional log-mel features using a 2048-sample Hann window, hop 1600, and non-centered windows. The shared per-ear causal convolutional stack has 64 channels and dilation rates $(1,2,4)$; the subsequent two-layer causal CNN has 128 channels and dilations $(1,2)$. Both use kernel size three and left padding. A layer-normalized projection, four-frame mean pooling, and a final linear projection produce acoustic tokens of width 768.

The final analysis window contributing to block $k$ ends at sample $(4k+3)1600+2048$. Thus its audio token becomes available at
\begin{equation}
 b_k=\frac{(4k+3)1600+2048}{48000},
\end{equation}
which is 9.33\,ms beyond the nominal four-frame boundary $4(k+1)/30$. This specifies the feature delay required by the factorization in Eq.~(\ref{eq:causal_factorization}). Section~\ref{sec:streaming_results} reports the subsequent motion-generation computation, starting from precomputed acoustic tokens. Measurement details are provided in Appendix~\ref{sec:streaming_efficiency}.

\paragraph{Spatial encoding.}
Spatial descriptors are normalized channel-wise using training-set statistics. The spatial MLP has widths $11\!\to\!768\!\to\!768$ with SiLU activation. Direction and distance increments are measured per latent step, rather than divided by elapsed time. Both increments are zero for the first block. Distance denominators are clamped from below at $10^{-8}$ for direction normalization and $10^{-6}$ for inverse distance and normalized approach. For SAM, we recover world-space source coordinates as the coordinatewise temporal median of $\mathbf{R}^{\mathrm{GT}}_t\boldsymbol{\ell}^{\mathrm{GT}}_t+\mathbf{p}^{\mathrm{GT}}_t$. At evaluation, the corresponding egocentric coordinates are updated from the generated root.

\paragraph{Guidance and sampling.}
For flow matching, we sample $\tau=\operatorname{sigmoid}(\xi)$ with $\xi\sim\mathcal{N}(0,1)$.
During training, the current acoustic and spatial conditions are jointly dropped with probability 0.1 by supplying zero-normalized inputs; the reaction-style embedding is retained. At inference, the conditional and null branches share the motion history and cached prefix, but differ in the current acoustic and spatial tokens. The guided velocity is
\begin{equation}
 v_\omega=v_{\mathrm{null}}+\omega(v_{\mathrm{cond}}-v_{\mathrm{null}}),
\end{equation}
where $\omega$ is the guidance scale; $v_{\mathrm{cond}}$ and $v_{\mathrm{null}}$ denote the respective branch predictions.
Only the conditional branch updates the persistent key--value cache. This guidance rule preserves past conditional states in both branches. Evaluation uses ten uniform Euler steps per generated latent. Training rollouts also use ten steps and unguided sampling.

\section{Training Sequence Construction}
\label{sec:training_sequences}

\paragraph{Two-forward training.}
Let $\widehat{\mathbf{z}}_i$ denote a detached first-pass denoising estimate. For full replacement, the two sequence passes are
\begin{align*}
 [\mathbf{z}^{\mathrm{GT}}_0,\mathbf{z}^{\mathrm{GT}}_1,\mathbf{z}^{\mathrm{GT}}_2,\ldots]
 &\xrightarrow{\text{first pass}}
 [\widehat{\mathbf{z}}_1,\widehat{\mathbf{z}}_2,\widehat{\mathbf{z}}_3,\ldots],\\
 [\mathbf{z}^{\mathrm{GT}}_0,\widehat{\mathbf{z}}_1,\widehat{\mathbf{z}}_2,\ldots]
 &\xrightarrow{\text{second pass}}\text{next-latent losses against GT}.
\end{align*}
Audio and spatial tokens are omitted for clarity. The first motion latent is retained, while first-pass predictions replace later inputs at corresponding positions. With updated SSL, the replaced sequence is decoded before the second pass to construct its spatial conditions. The second pass is causally masked, but its outputs do not become later inputs within that pass.

For normalized training progress $u\in[0,1]$, replacement schedules are
\begin{equation}
 p_{0\to1}(u)=\tfrac12(1-\cos\pi u),\quad
 p_{1\to0}(u)=\tfrac12(1+\cos\pi u),\quad
 p_{\mathrm{constant}}(u)=1.
\end{equation}
A subset of input positions is randomly selected according to the replacement fraction. These schedules alter motion-input replacement rather than switching between spatial-conditioning protocols.

\paragraph{Rehearsal initialization and recursion.}
For each sequence with $K$ valid latents, the start $s$ is sampled uniformly from $\{1,\ldots,K-H-1\}$, with $H=16$ by default. The Online result in Table~\ref{tab:reference_results} uses $H=32$. The rollout seed is
\begin{equation}
 \overline{\mathbf{z}}_s=\operatorname{sg}\!\left[
 \mathbf{u}^{(s)}_\tau+(1-\tau)
 v_\theta(\mathbf{u}^{(s)}_\tau,\tau;\mathbf{h}^{\mathrm{GT}}_{s-1})\right],
 \label{eq:rollout_seed}
\end{equation}
where $\mathbf{u}^{(s)}_\tau$ is an interpolation between noise and the GT target $\mathbf{z}^{\mathrm{GT}}_s$, and $\operatorname{sg}$ denotes stop-gradient. Subsequent latents are sampled from fresh Gaussian noise and appended recursively, as in Eq.~(\ref{eq:rehearsal_recursion}).
The decoder receives the pre-window GT prefix followed by these generated latents. With online spatial updates, the decoded root determines the next spatial descriptor after each append; the GT-SSL variant retains recorded conditions. Each transition is supervised with the next GT latent, independently of the sampled latent used to extend the history.

The acoustic features are cropped at frame $rs$ and re-encoded. Acoustic convolutional context and the transformer's key--value cache are reset at this boundary, while the decoder retains its GT prefix. Each rollout contains $H$ transitions, producing $4H$ frames. Gradients are stopped through the sampled latents and decoded geometry, but retained through transformer states and conditioning encoders in the supervised objective.

\paragraph{Rollout-only ablations.}
Both variants optimize only the rollout loss and retain the stop-gradient initialization in Eq.~(\ref{eq:rollout_seed}); the teacher-forced prediction pass supplies the seed but contributes no training loss. Rollout Only uses the same independently sampled starts and reset acoustic and transformer contexts as rehearsal. Rollout Only with History instead builds a transformer cache from the GT motion and conditioning prefix before $s$, and takes acoustic conditions from the full causal encoding. Gradients from the rollout loss propagate through this prefix cache. Both variants use $H=16$ and ten sampling steps per latent.

%% file: sections/appendix_visual_results.tex
\section{Additional Visual Results and Physics-Based Simulation}
\label{sec:additional_visual_results}

The supplementary material includes HTML and videos of four comparisons with MOSPA, ten single-sound examples, and a continuous sequence spanning three sound events. To view the videos in the accompanying HTML page, open \texttt{index.html} after extracting the supplementary package. The single-sound and continuous examples pair human motion visualizations with humanoid physics simulations, synchronized with stereo audio.

\noindent\begin{minipage}{\linewidth}
 \centering
 \input{figures/appendix_paired_sequence}
 
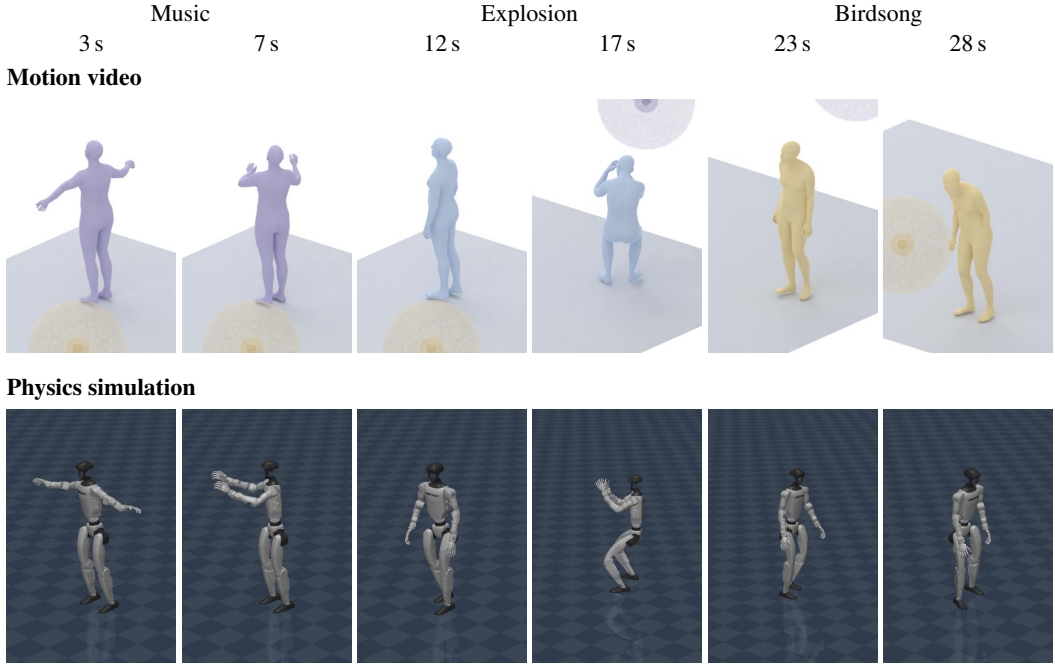
\captionof{figure}{\textbf{Human reactions and humanoid tracking in physics simulation.} Aligned snapshots of the continuous Music $\rightarrow$ Explosion $\rightarrow$ Birdsong sequence show human motion (top) and G1 simulation (bottom). Robot tracking is described in Appendix~\ref{sec:physics_simulation}.}
 \label{fig:paired_motion_simulation}
\end{minipage}

\subsection{Reactions to Different Sounds}
\label{sec:additional_sound_reactions}

\begin{figure}[p]
 \centering
 \input{figures/appendix_reactions}
 \caption{\textbf{Additional reactions under online spatial feedback.} Original generated motions for six sounds, with poses ordered chronologically. Purple markers denote sound sources; clip identifiers match the supplementary videos.}
 \label{fig:additional_reactions}
\end{figure}
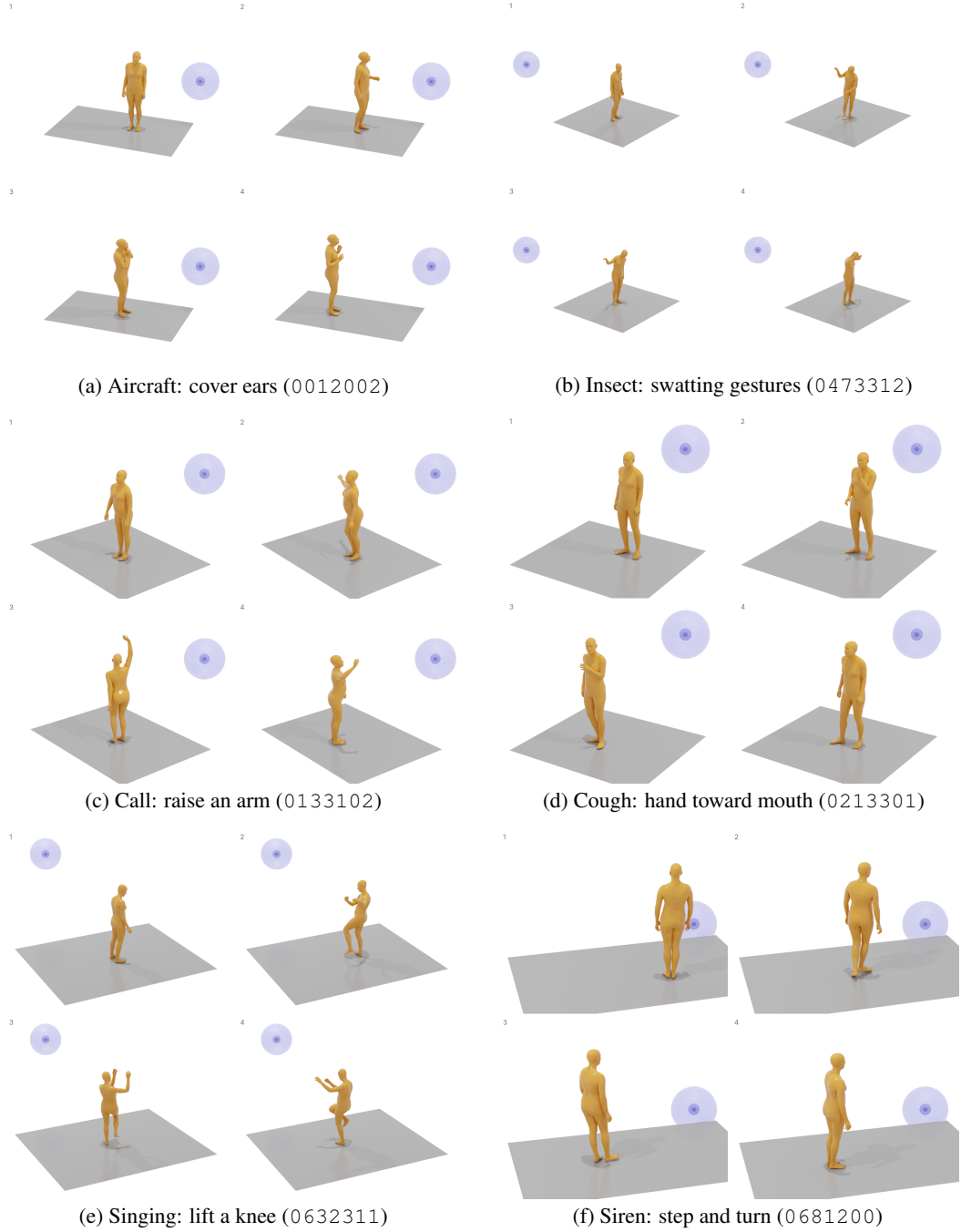

Across diverse sounds, Sonic-AMG combines recognizable gestures with changes in body orientation and position. Figures~\ref{fig:additional_reactions} and~\ref{fig:additional_reactions_more} show all ten single-sound examples from Part 3 of the supplementary HTML, including hands raised toward the ears, an arm raised in response to a call, and a hand brought toward the mouth. The full videos show how these gestures develop alongside stepping and body turns. All examples use the feedback-trained model with online spatial updates.

\clearpage
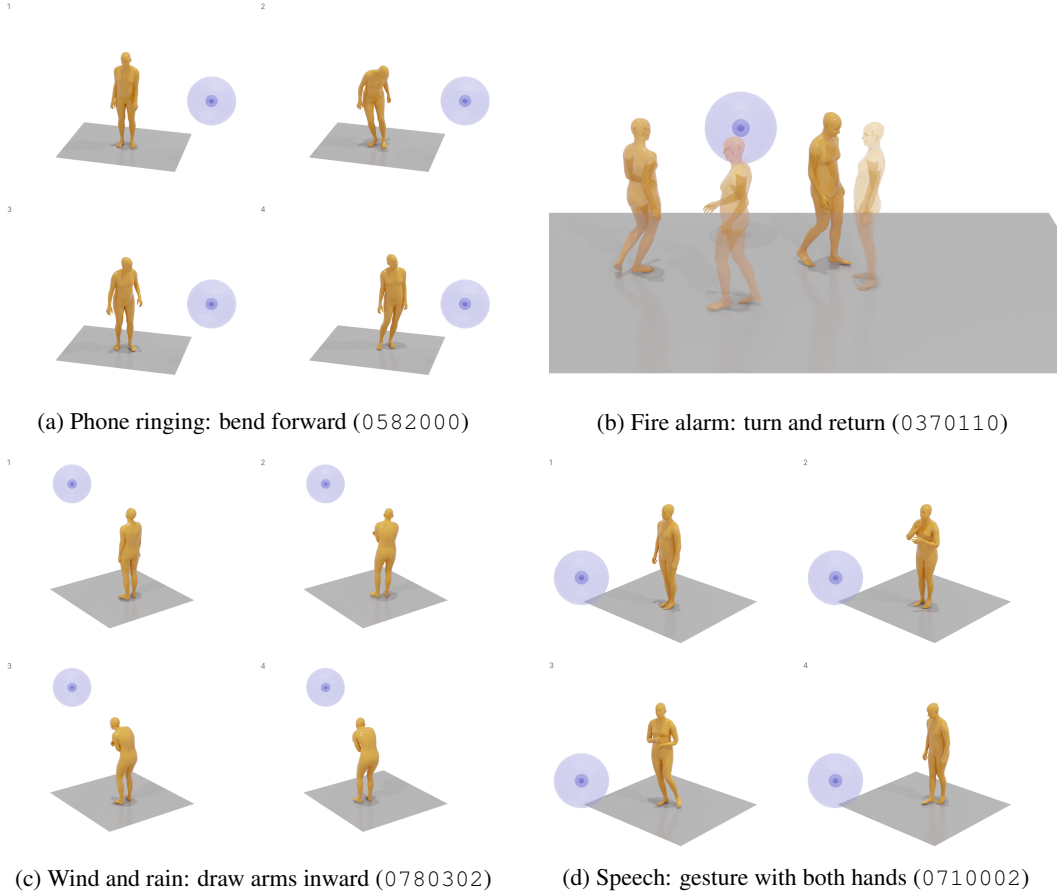
\begin{figure}[!t]
 \centering
 \input{figures/appendix_reactions_more}
 \caption{\textbf{Additional reactions under online spatial feedback (continued).}}
 \label{fig:additional_reactions_more}
\end{figure}

\subsection{Continuous Reactions across Sound Events}
\label{sec:continuous_details}

Sonic-AMG sustains coherent reactions as sound content and source location change, without a new ground-truth motion initialization at event transitions. The continuous example spans 30.13\,s and combines three SAM audio events---Music $\rightarrow$ Explosion $\rightarrow$ Birdsong---at different source locations (Table~\ref{tab:continuous_sequence_setup}). Generation begins with four GT motion frames; later events continue from the generated body state.

\begin{table}[htbp]
 \centering
 \caption{\textbf{Continuous demonstration setup.} Source locations change at event boundaries. Intervals refer to the generated 30\,Hz sequence; style identifiers follow SAM.}
 \label{tab:continuous_sequence_setup}
 \small
 \begin{tabular}{llcr}
  \toprule
  Sound & SAM clip & Style & Interval (s) \\
  \midrule
  Music & \texttt{0541212} & 2 & $[0,10.00)$ \\
  Explosion & \texttt{0330202} & 2 & $[10.00,20.27)$ \\
  Birdsong & \texttt{0113311} & 1 & $[20.27,30.13)$ \\
  \bottomrule
 \end{tabular}
\end{table}

At each transition, generation resumes from the generated motion history in a root-aligned frame. Spatial feedback updates the active source's body-relative coordinates, while previously emitted frames remain fixed.

\subsection{Execution in a Humanoid Physics Simulation}
\label{sec:physics_simulation}

Generated reactions can be tracked by a pretrained humanoid controller without additional policy training. We retarget the human motions to a Unitree G1 humanoid using General Motion Retargeting (GMR)~\citep{araujo2026gmr} and track the references offline with the released pretrained GEAR-SONIC controller~\citep{luo2026sonic} in MuJoCo~\citep{todorov2012mujoco}. References are interpolated from 30 to 50\,Hz, with physics integrated at 1,000\,Hz. The simulation uses gravity, ground contact, and PD torques clipped to actuator limits. After initialization, the free root evolves through physics integration.

Figure~\ref{fig:paired_motion_simulation} pairs human reactions with their simulated humanoid execution. The robot tracks the references through all three sound events in a single uninterrupted simulation.

%% file: figures/appendix_paired_sequence.tex
\begingroup
\setlength{\tabcolsep}{1.2pt}
\begin{tabular}{@{}cccccc@{}}
\multicolumn{2}{c}{\small Music} & \multicolumn{2}{c}{\small Explosion} & \multicolumn{2}{c}{\small Birdsong} \\
\footnotesize 3\,s & \footnotesize 7\,s & \footnotesize 12\,s & \footnotesize 17\,s & \footnotesize 23\,s & \footnotesize 28\,s \\[2pt]
\multicolumn{6}{@{}l}{\small\textbf{Motion video}} \\[2pt]
\includegraphics[width=.16\linewidth,trim={258bp 380bp 442bp 50bp},clip]{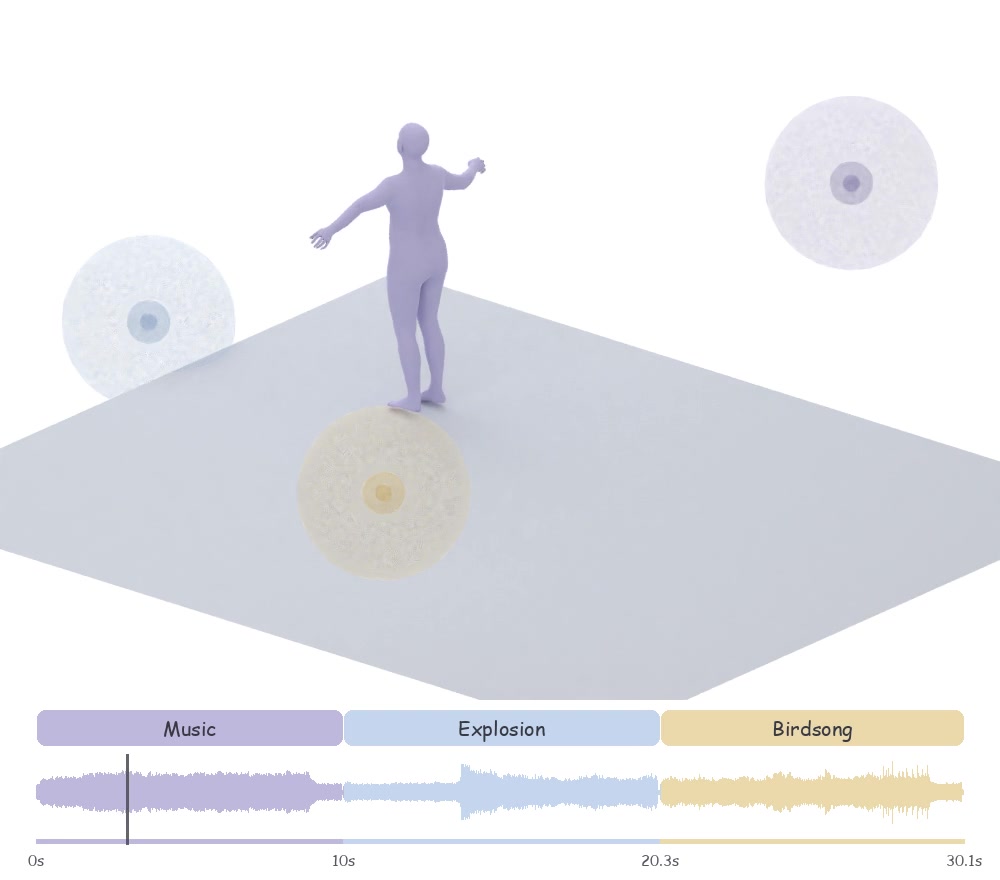} &
 \includegraphics[width=.16\linewidth,trim={260bp 380bp 440bp 50bp},clip]{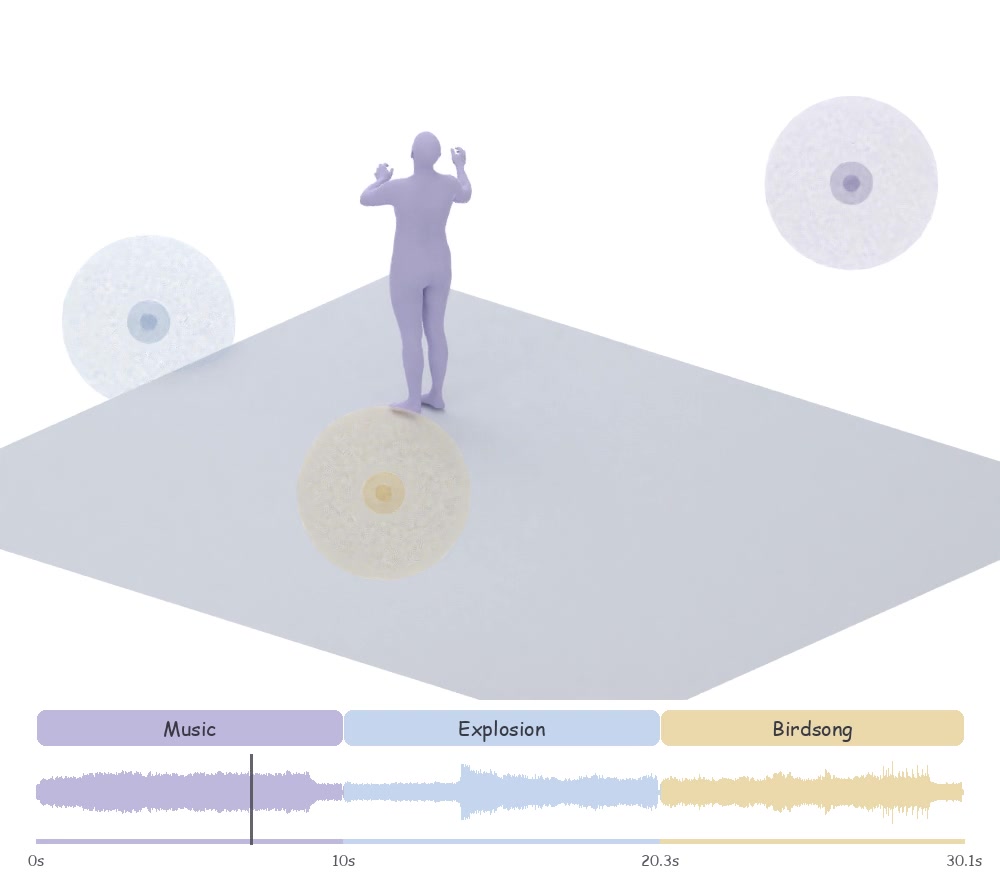} &
 \includegraphics[width=.16\linewidth,trim={253bp 380bp 447bp 50bp},clip]{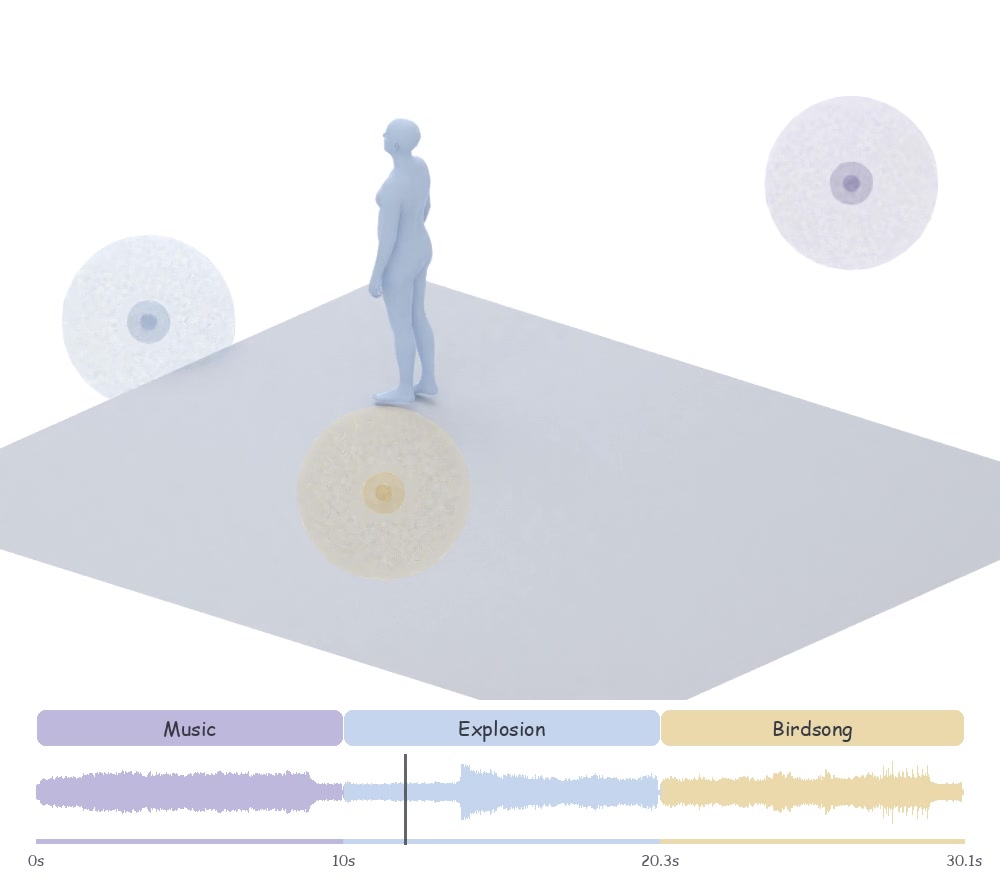} &
 \includegraphics[width=.16\linewidth,trim={650bp 250bp 50bp 180bp},clip]{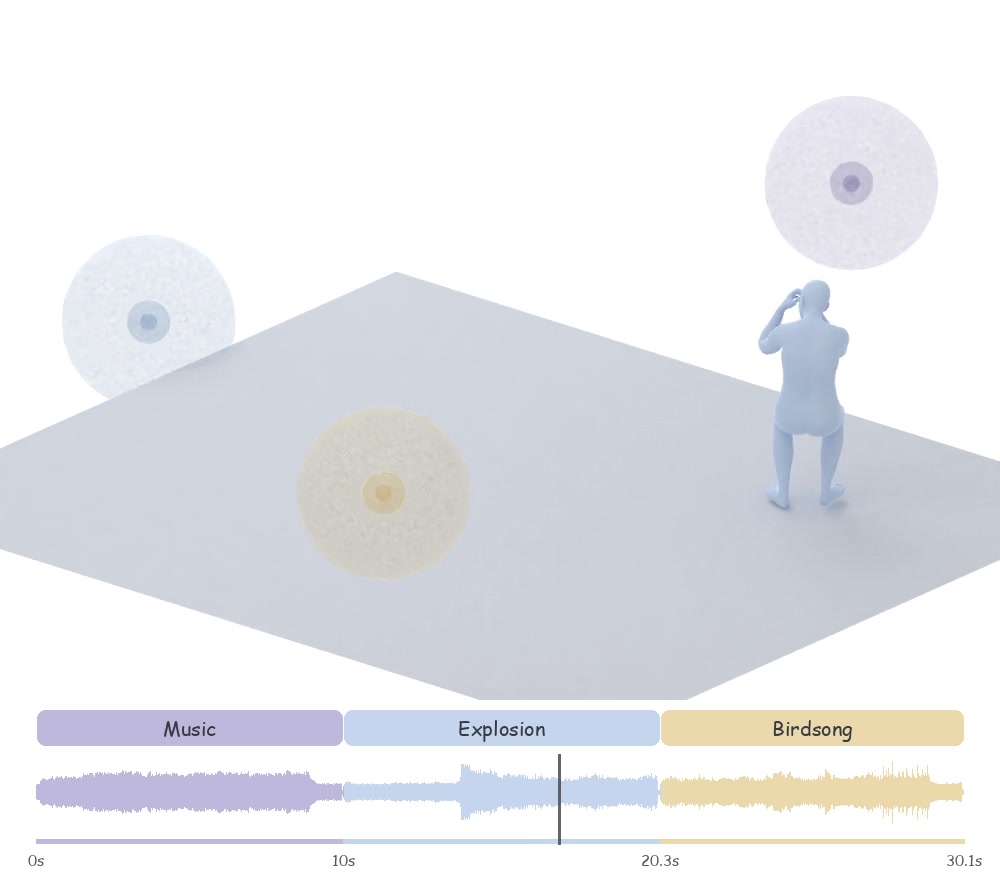} &
 \includegraphics[width=.16\linewidth,trim={590bp 200bp 110bp 230bp},clip]{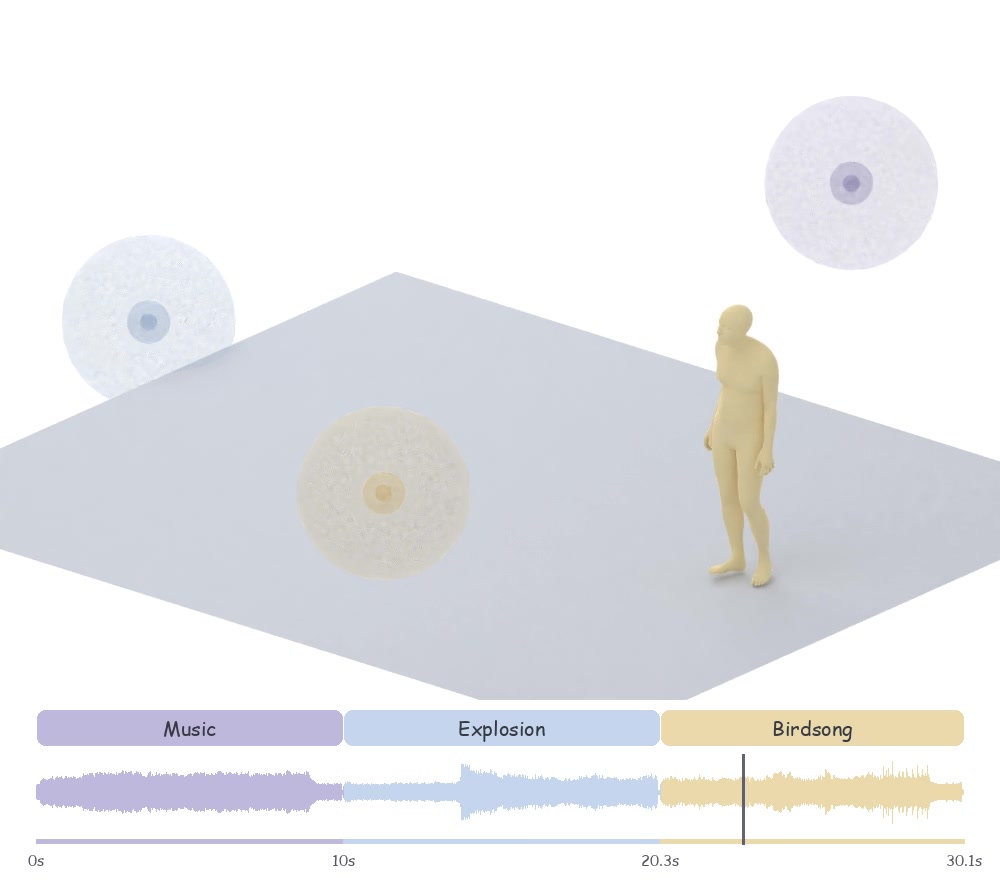} &
 \includegraphics[width=.16\linewidth,trim={350bp 195bp 350bp 235bp},clip]{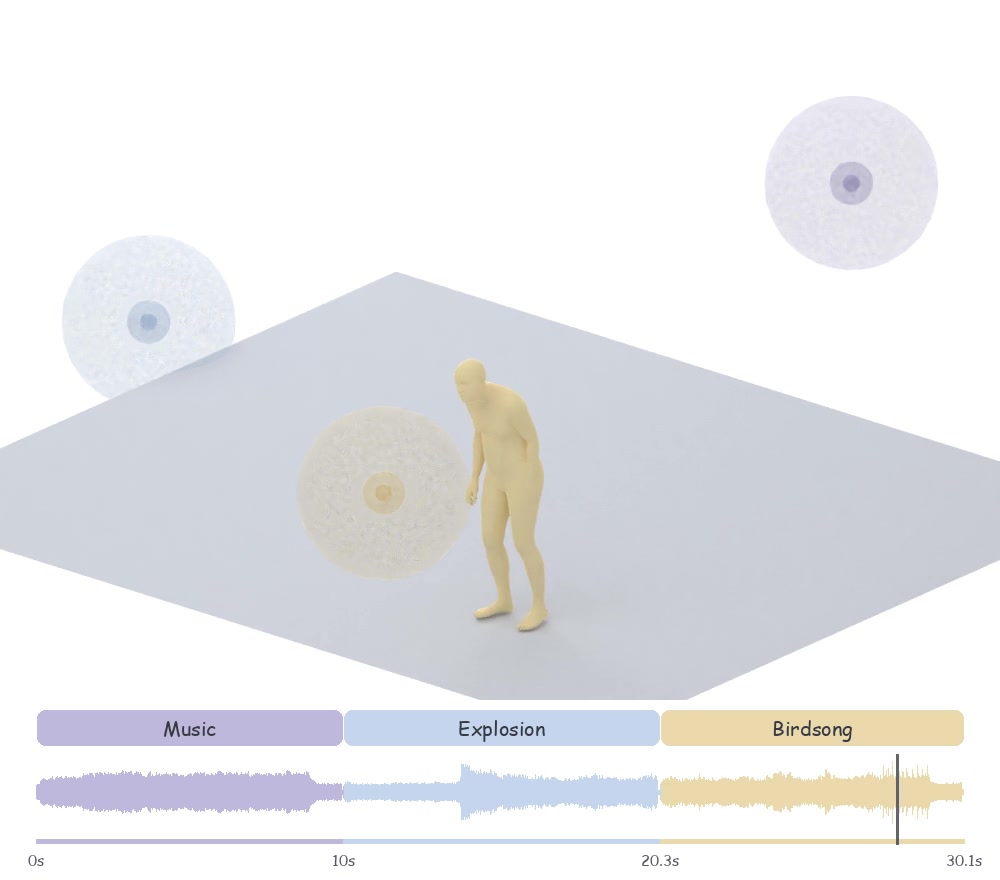} \\[5pt]
\multicolumn{6}{@{}l}{\small\textbf{Physics simulation}} \\[2pt]
\includegraphics[width=.16\linewidth,trim={408bp 420bp 1152bp 480bp},clip]{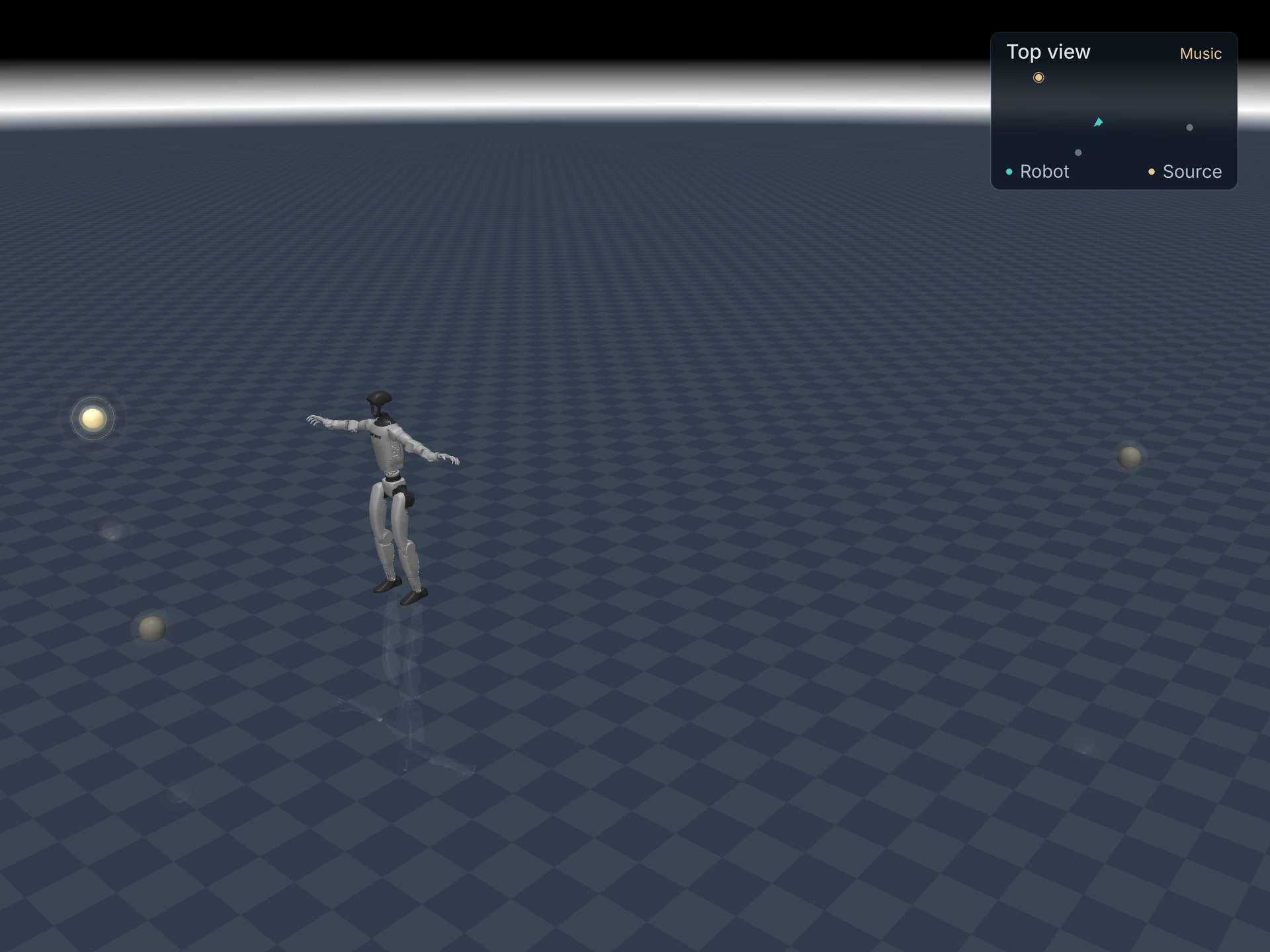} &
 \includegraphics[width=.16\linewidth,trim={410bp 420bp 1150bp 480bp},clip]{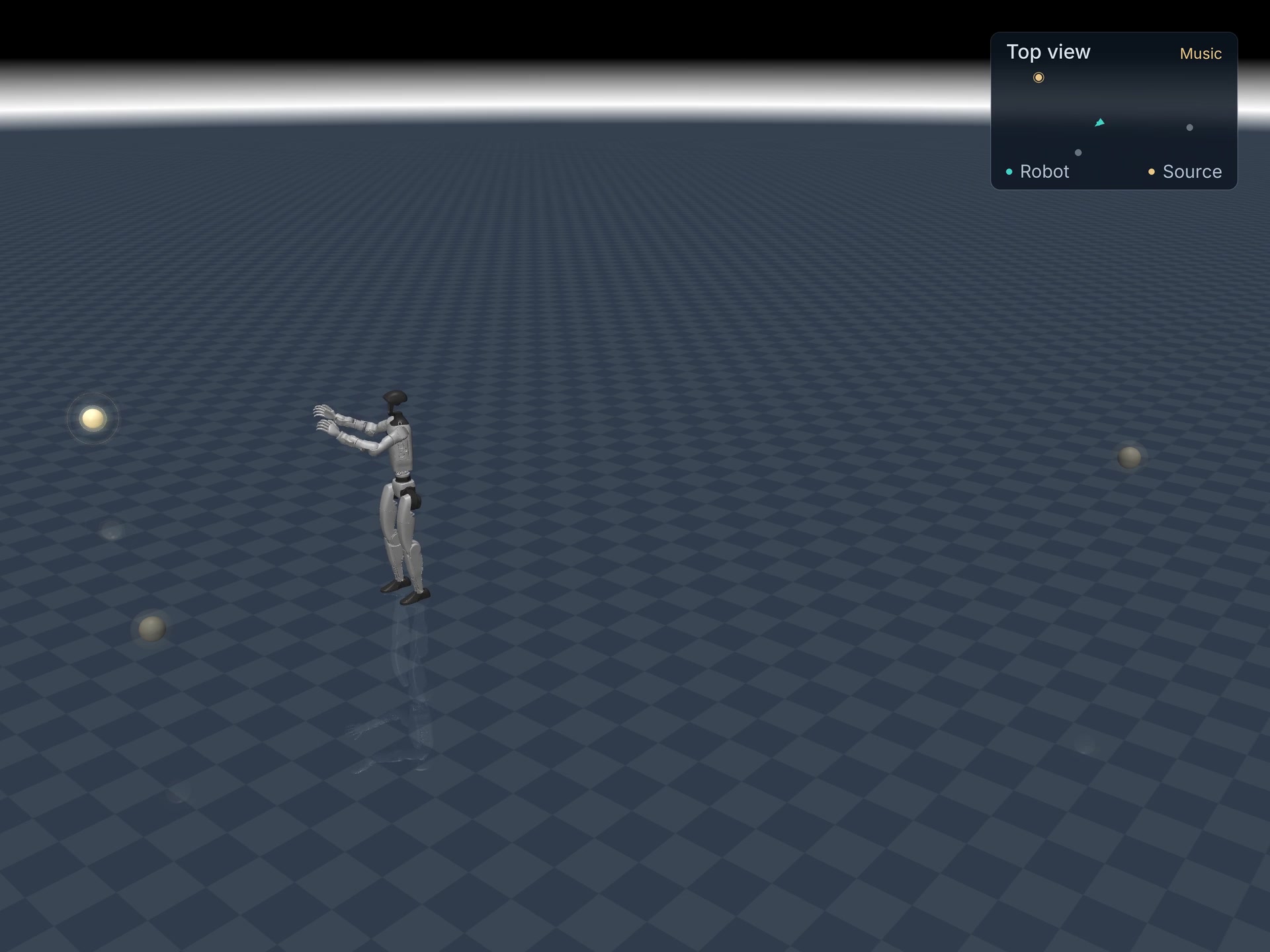} &
 \includegraphics[width=.16\linewidth,trim={430bp 420bp 1130bp 480bp},clip]{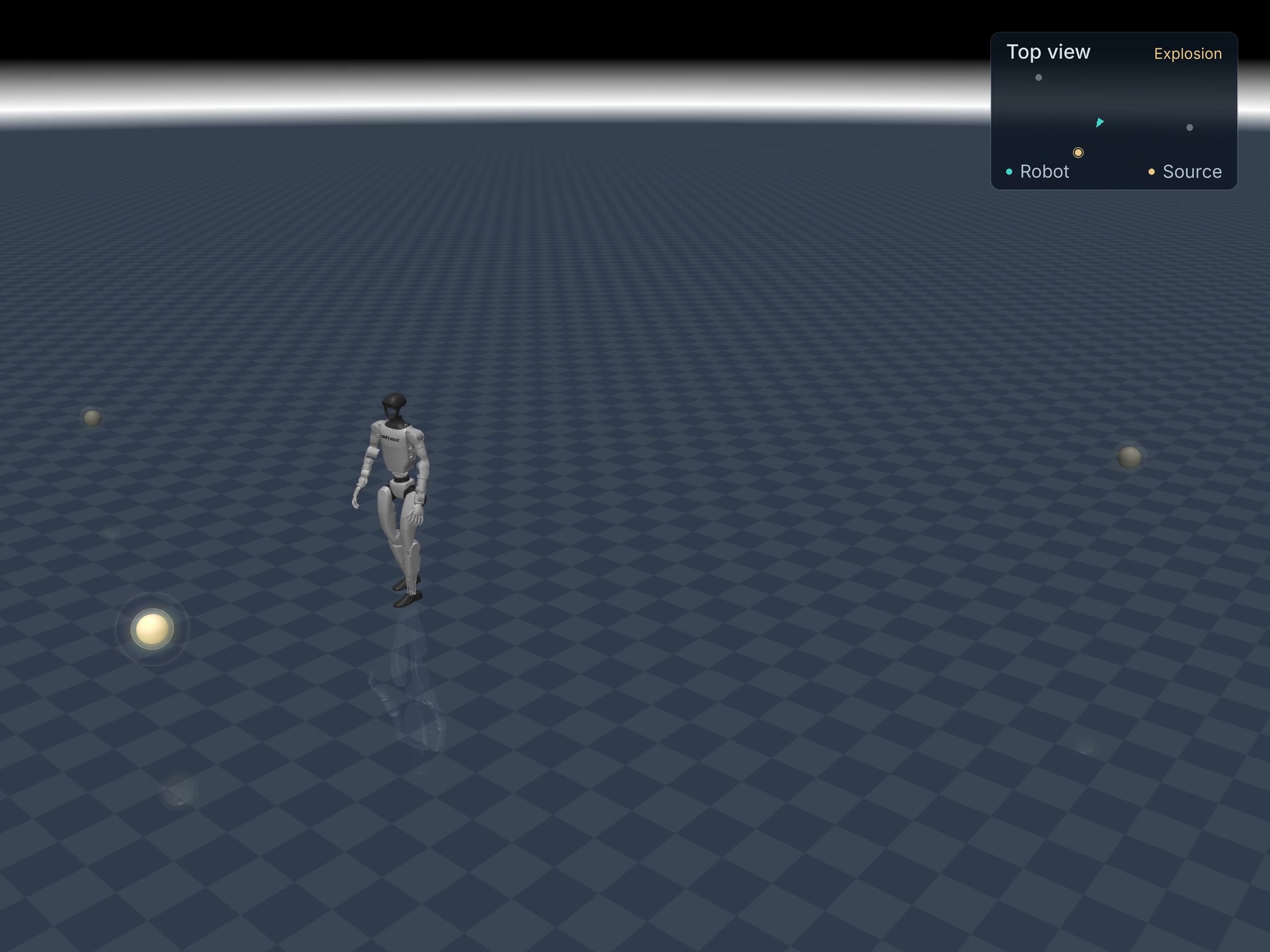} &
 \includegraphics[width=.16\linewidth,trim={700bp 490bp 860bp 410bp},clip]{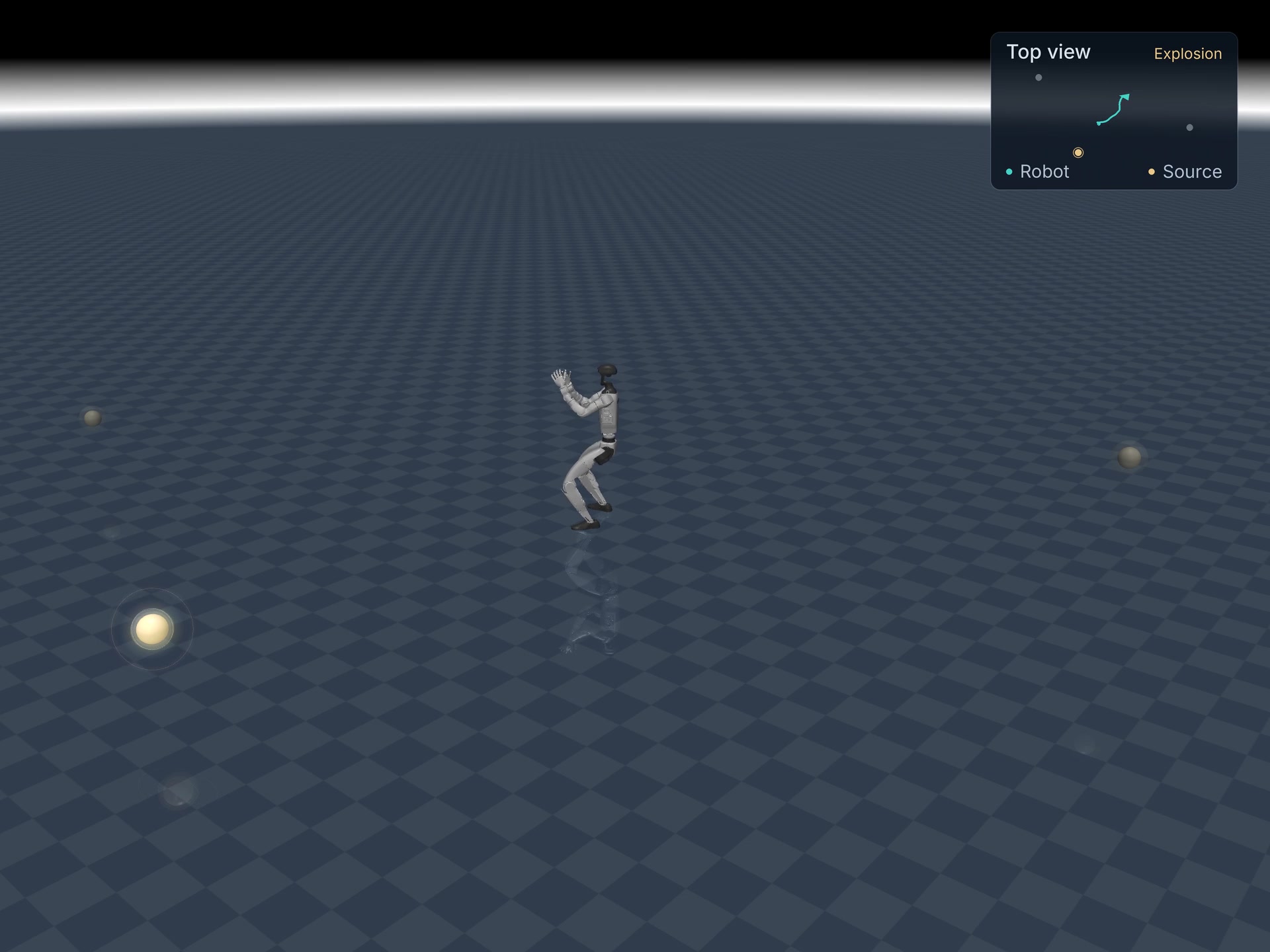} &
 \includegraphics[width=.16\linewidth,trim={750bp 500bp 810bp 400bp},clip]{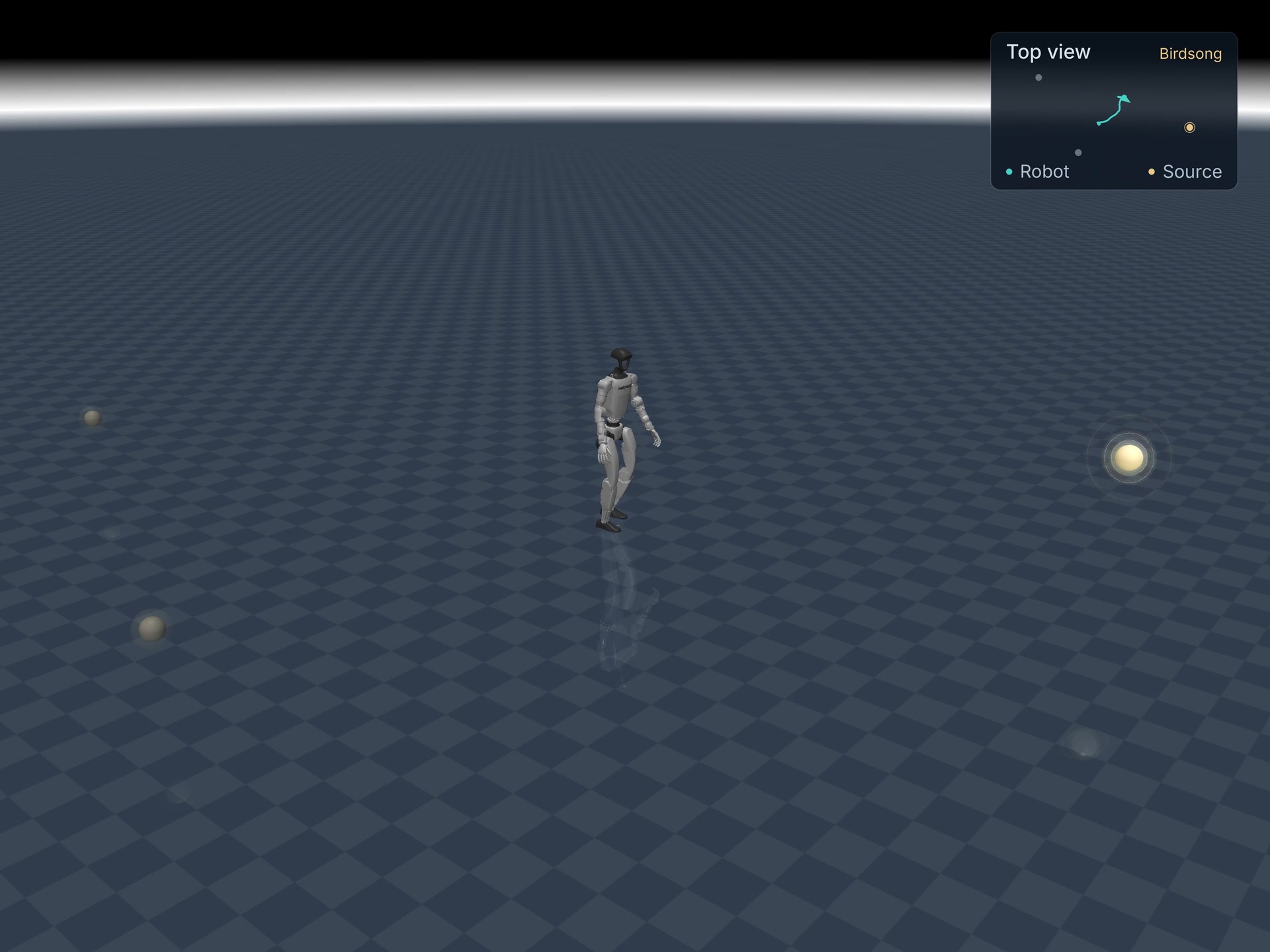} &
 \includegraphics[width=.16\linewidth,trim={900bp 450bp 660bp 450bp},clip]{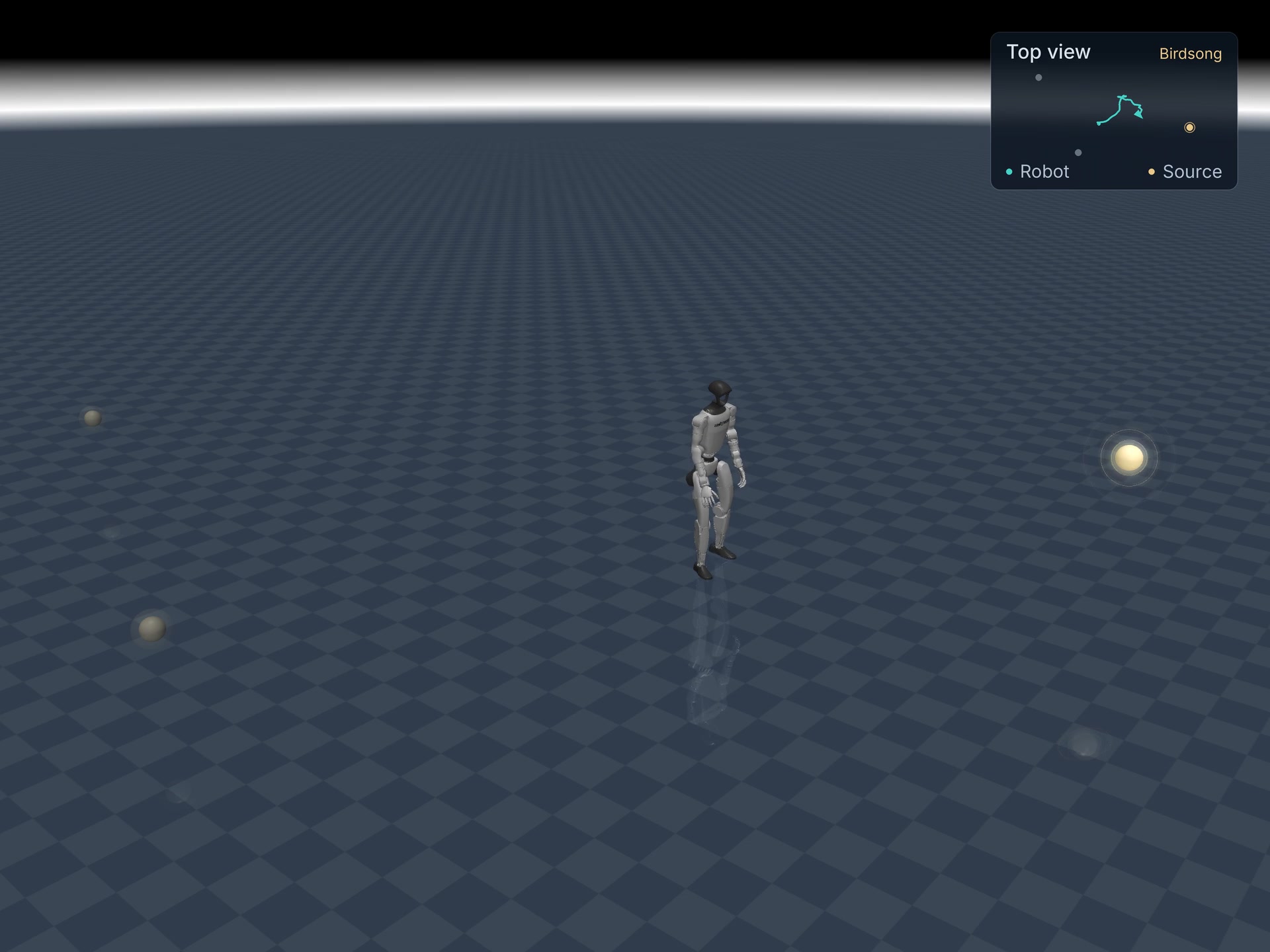} \\[5pt]
\end{tabular}
\endgroup

%% file: figures/appendix_reactions.tex
\begin{minipage}{.48\linewidth}
 \centering
 \includegraphics[width=\linewidth]{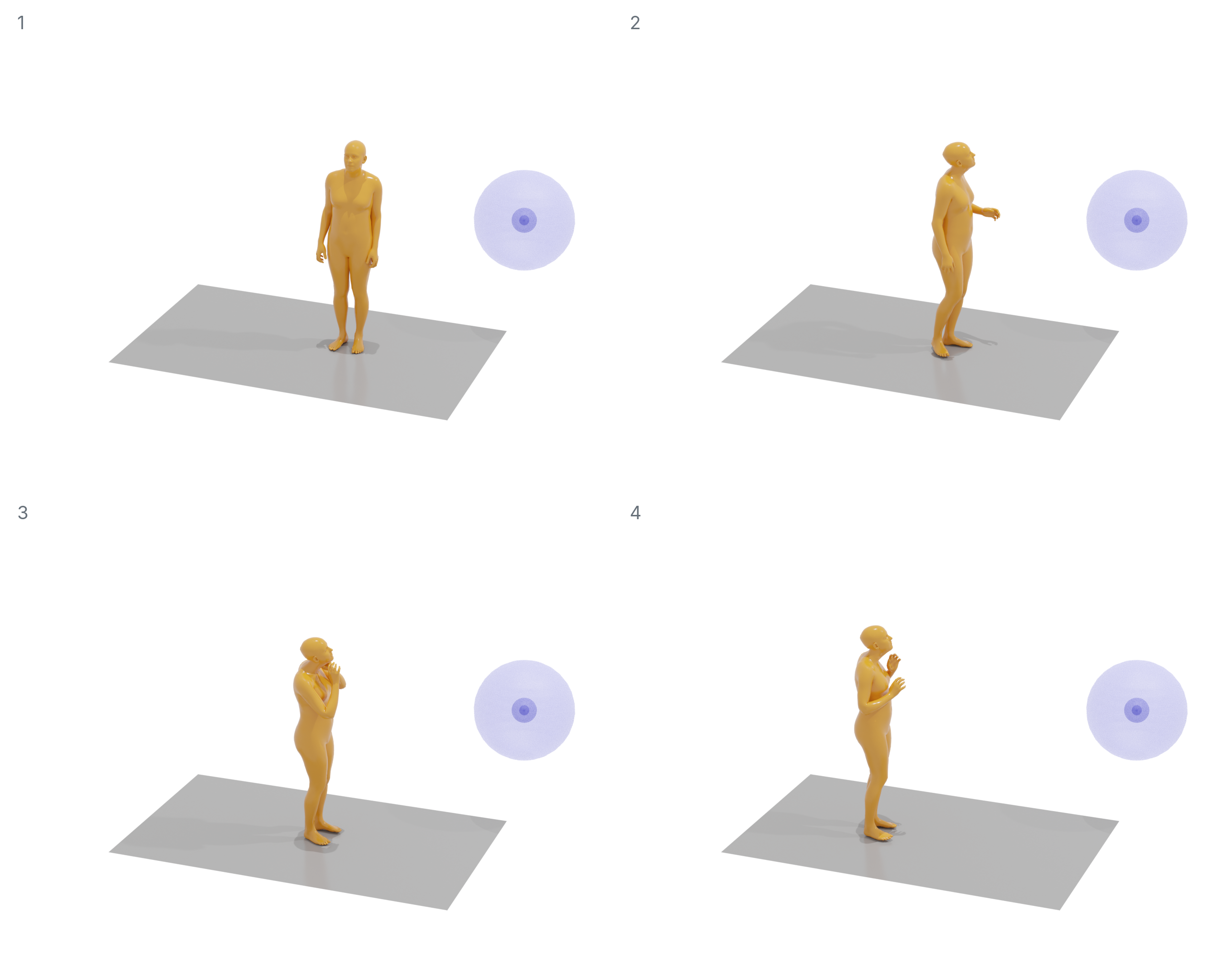}
 \small (a) Aircraft: cover ears (\texttt{0012002})
\end{minipage}\hfill
\begin{minipage}{.48\linewidth}
 \centering
 \includegraphics[width=\linewidth]{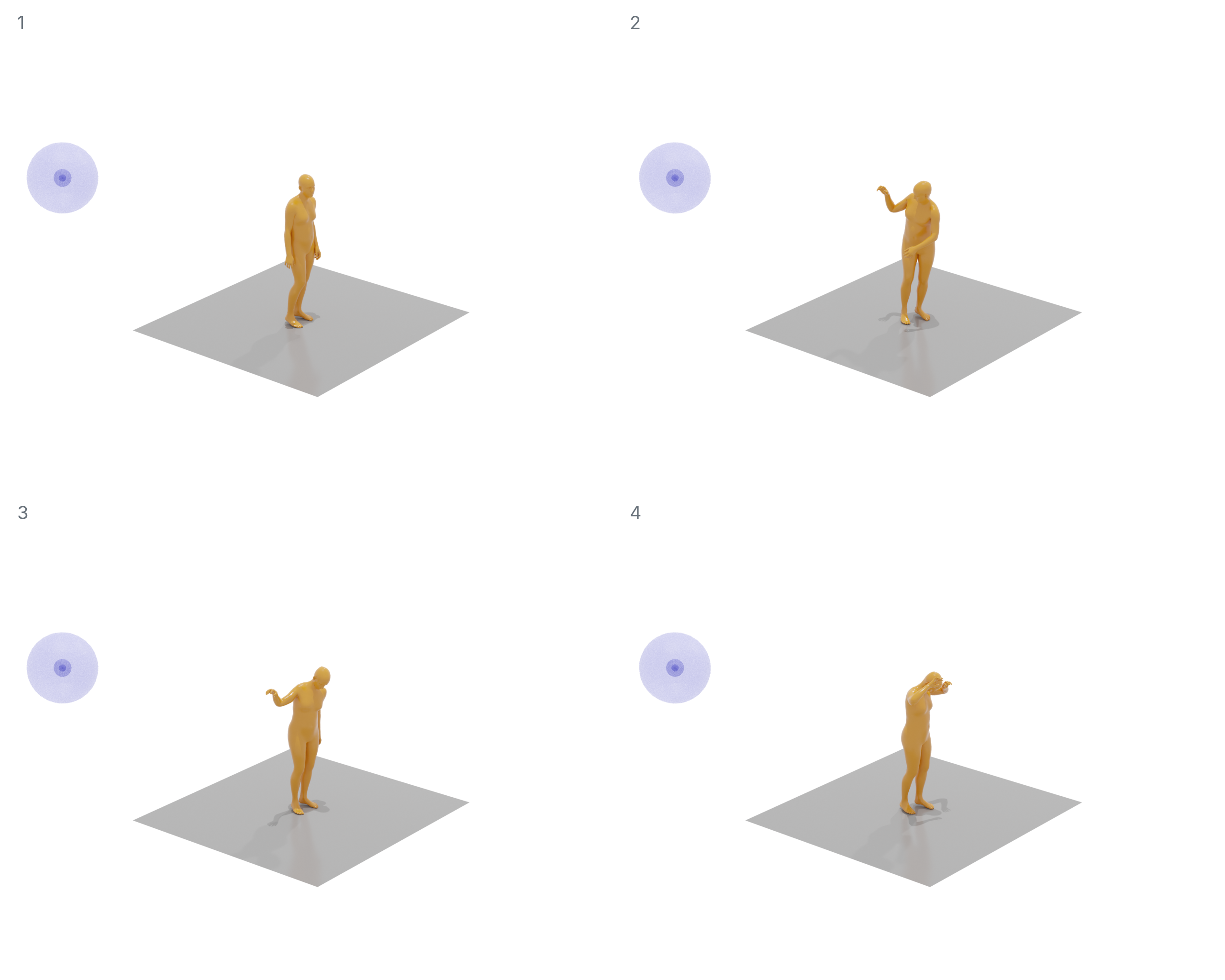}
 \small (b) Insect: swatting gestures (\texttt{0473312})
\end{minipage}

\medskip
\begin{minipage}{.48\linewidth}
 \centering
 \includegraphics[width=\linewidth]{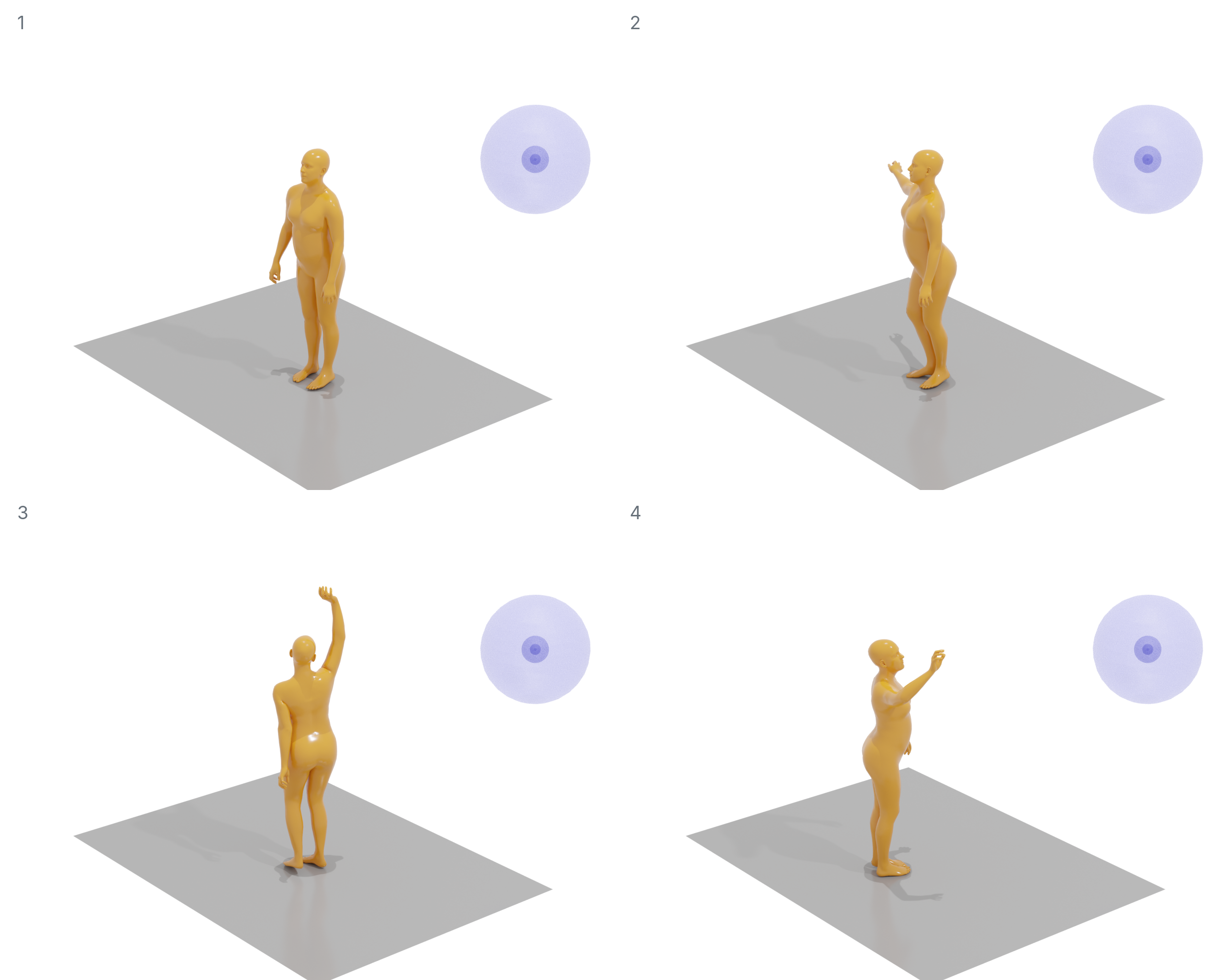}
 \small (c) Call: raise an arm (\texttt{0133102})
\end{minipage}\hfill
\begin{minipage}{.48\linewidth}
 \centering
 \includegraphics[width=\linewidth]{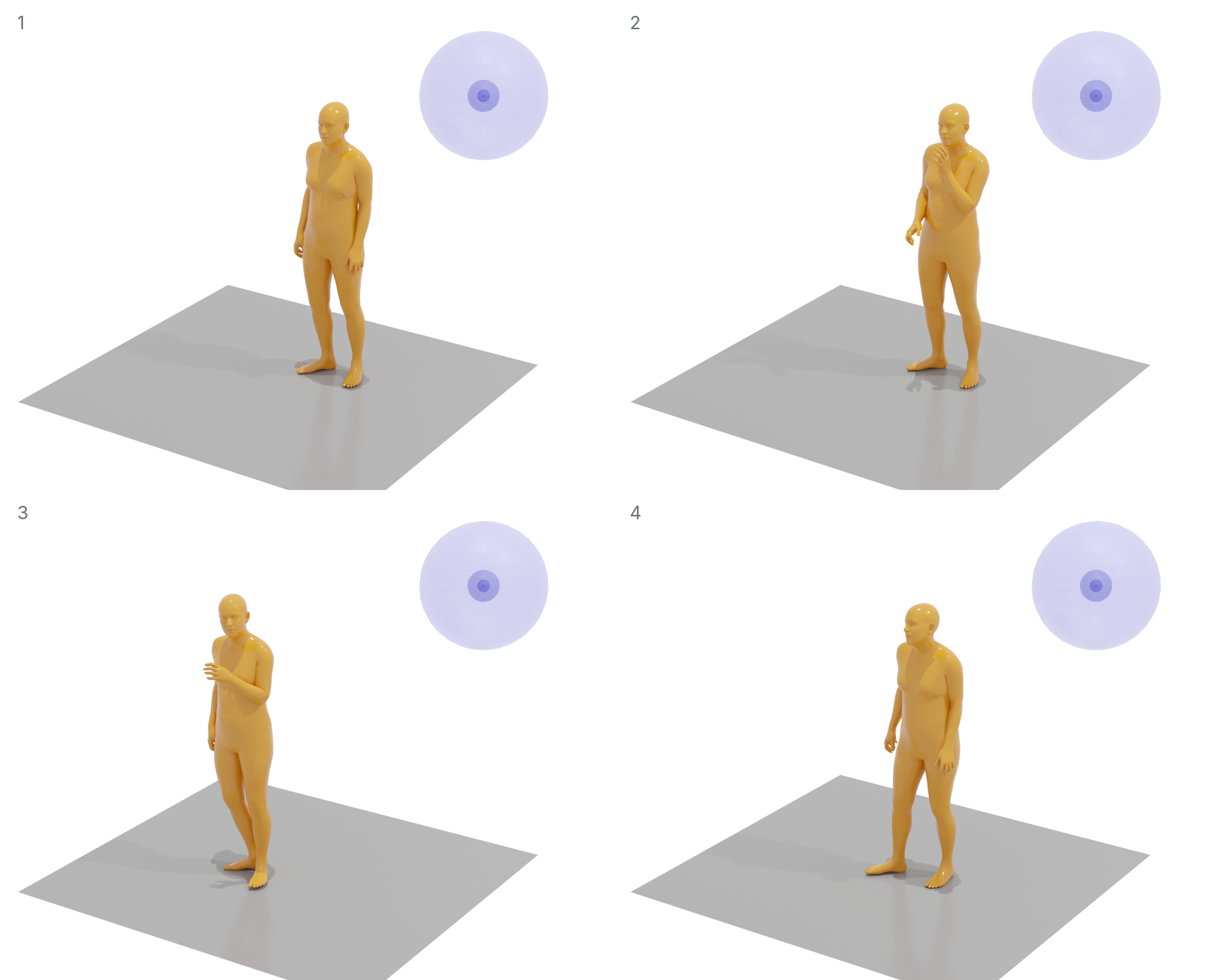}
 \small (d) Cough: hand toward mouth (\texttt{0213301})
\end{minipage}

\medskip
\begin{minipage}{.48\linewidth}
 \centering
 \includegraphics[width=\linewidth]{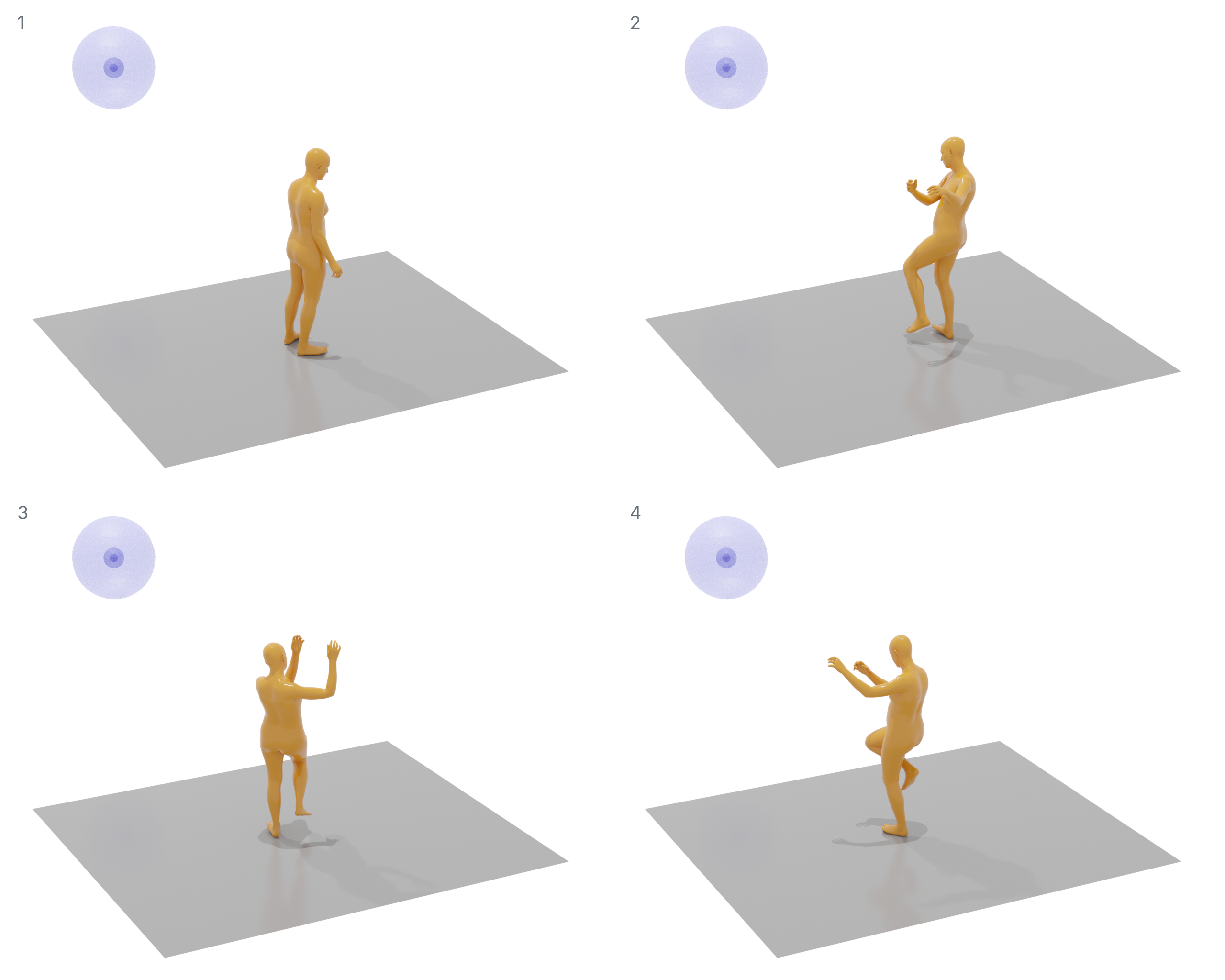}
 \small (e) Singing: lift a knee (\texttt{0632311})
\end{minipage}\hfill
\begin{minipage}{.48\linewidth}
 \centering
 \includegraphics[width=\linewidth]{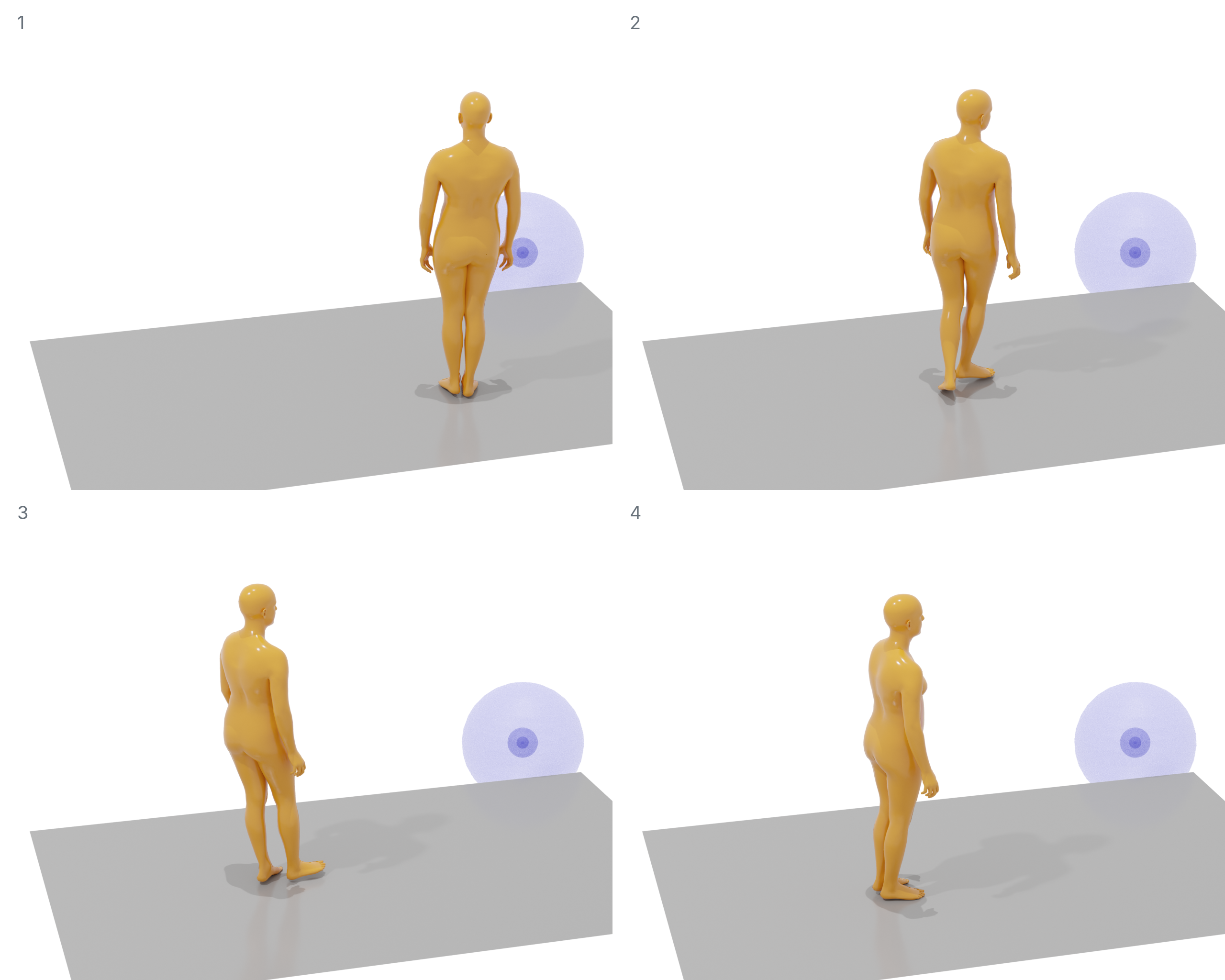}
 \small (f) Siren: step and turn (\texttt{0681200})
\end{minipage}

%% file: figures/appendix_reactions_more.tex
\begin{minipage}{.48\linewidth}
 \centering
 \includegraphics[width=\linewidth]{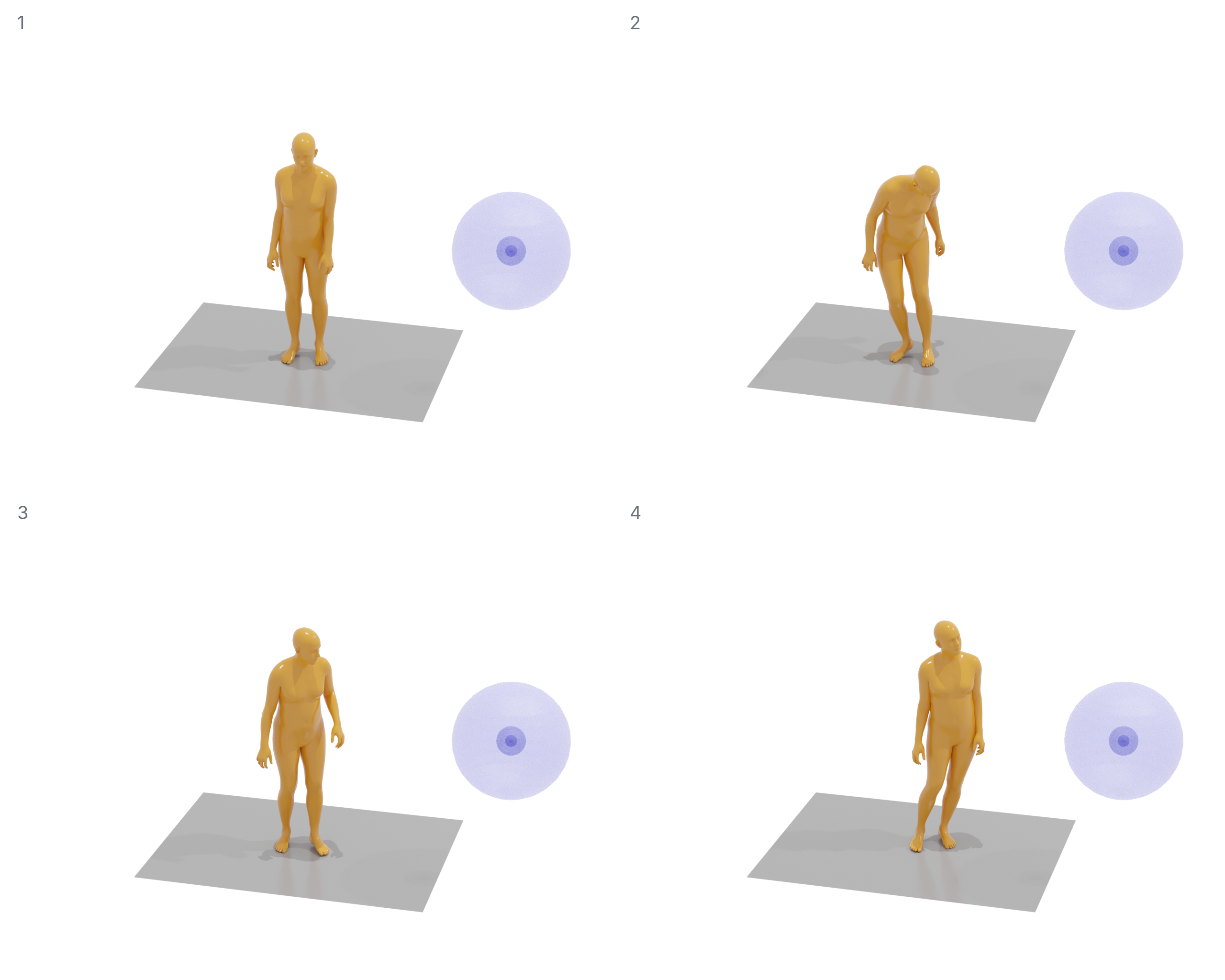}
 \small (a) Phone ringing: bend forward (\texttt{0582000})
\end{minipage}\hfill
\begin{minipage}{.48\linewidth}
 \centering
 \includegraphics[width=\linewidth]{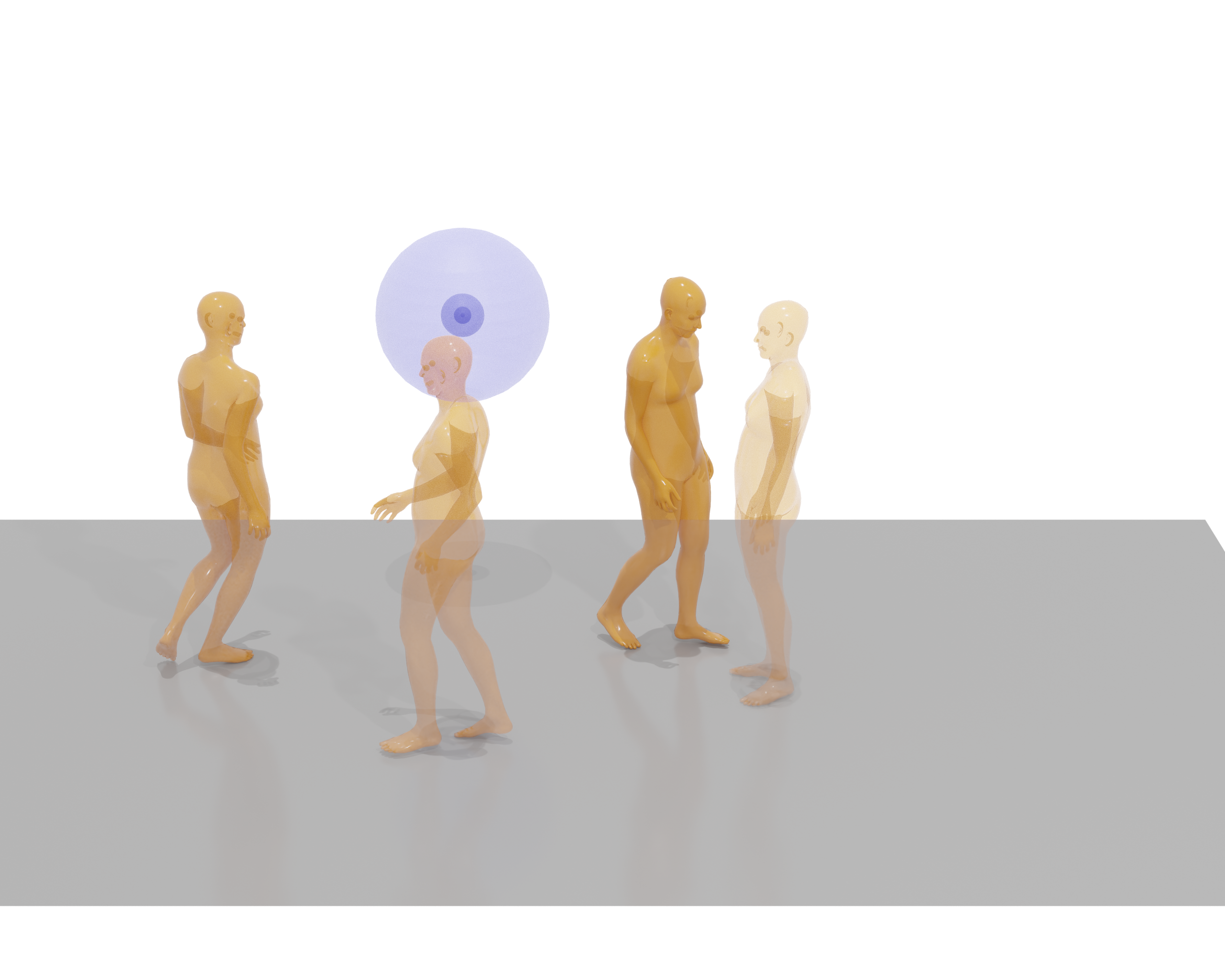}
 \small (b) Fire alarm: turn and return (\texttt{0370110})
\end{minipage}

\medskip
\begin{minipage}{.48\linewidth}
 \centering
 \includegraphics[width=\linewidth]{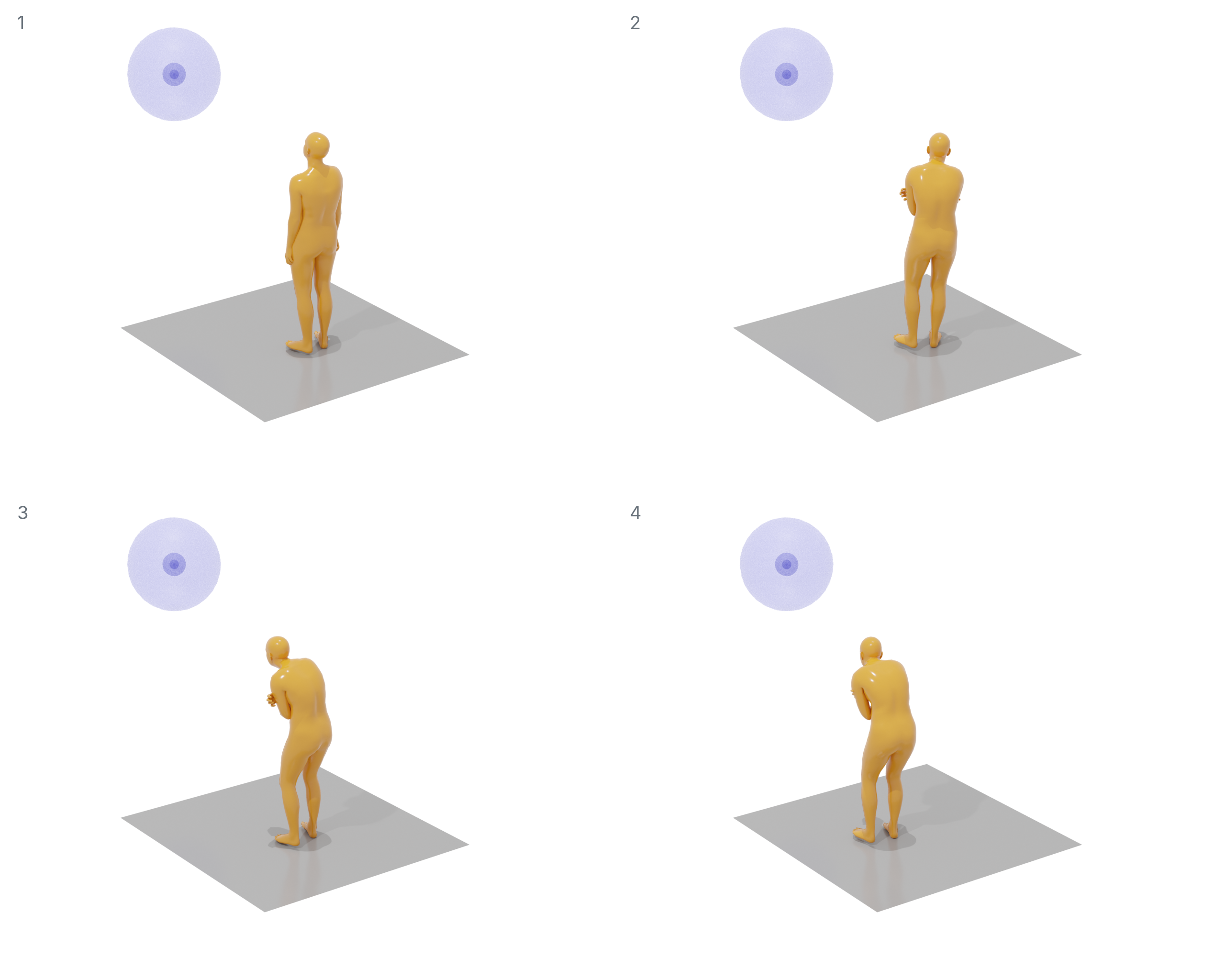}
 \small (c) Wind and rain: draw arms inward (\texttt{0780302})
\end{minipage}\hfill
\begin{minipage}{.48\linewidth}
 \centering
 \includegraphics[width=\linewidth]{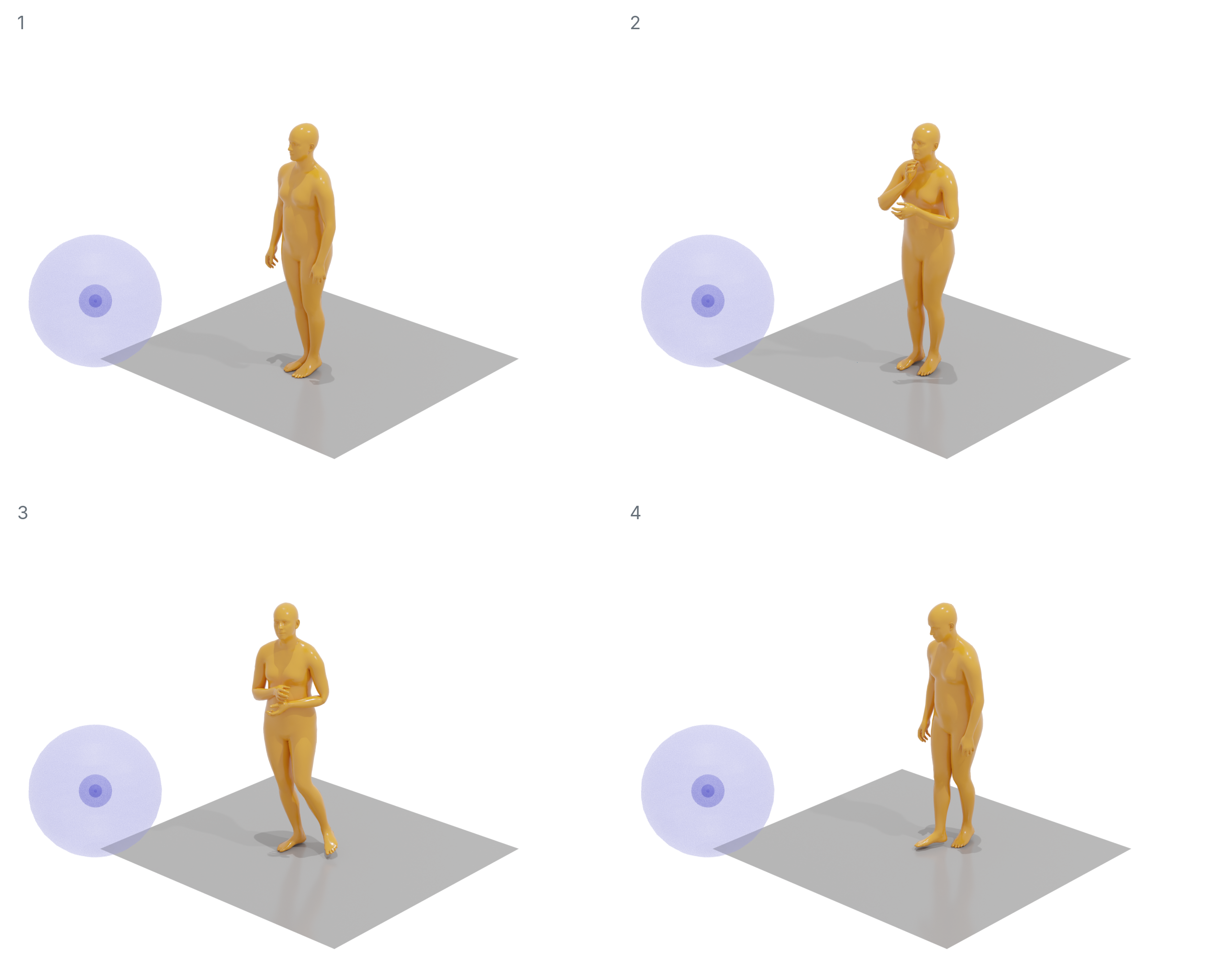}
 \small (d) Speech: gesture with both hands (\texttt{0710002})
\end{minipage}

%% file: sections/appendix_streaming.tex
\section{Streaming Timing Protocols}
\label{sec:streaming_efficiency}
\label{sec:streaming_benchmark}

This appendix details the computation measurements and initial response estimates reported in Section~\ref{sec:streaming_results} and Table~\ref{tab:streaming_results}.

\paragraph{Timing protocol.}
We measure one 700-step run per GPU and spatial protocol, discarding the first 20 steps as warm-up. Generation starts from a zero motion latent. The run measures computation over 93.3\,s of additional motion. The KV cache retains the full history within a configured context of 4,096 token positions. We use FP32 tensors with TF32 enabled where supported and synchronize the GPU around each wall-clock measurement. The GT and Online runs use the GT-SSL and feedback-trained rehearsal models, respectively.

\paragraph{Measured operations.}
At batch size one, each step includes both guidance branches, ten Euler sampling steps, and the persistent KV-cache update. GT spatial tokens are precomputed. Online timing additionally includes decoding the generated latent prefix and computing spatial tokens from its last two poses; quality evaluation uses four-frame pooled coordinates (Section~\ref{sec:online_conditions}). We time the motion-generation stage from precomputed acoustic tokens, excluding audio preprocessing and encoding, initialization, rendering, and output delivery. The acoustic frontend and feature-availability delay are specified in Appendix~\ref{sec:implementation}.

\paragraph{Per-step computation.}
Online feedback increases mean step time from 41.2 to 50.2\,ms on RTX A6000 and from 102.5 to 118.0\,ms on TITAN RTX. The corresponding online real-time factors are 0.376 and 0.885, with $\mathrm{RTF}=\overline{t}_{\mathrm{step}}/(4/30\,\mathrm{s})$, where $\overline{t}_{\mathrm{step}}$ denotes mean per-step computation time.

\paragraph{MOSPA timing protocol.}
MOSPA jointly denoises a complete 240-frame sequence using the full audio window~\citep{xu2025mospa}, so its generation protocol does not support causal blockwise streaming. We time its released sampler on the same two GPUs: batch size one, the published 1{,}000-step denoising schedule, and five timed blocks after one warm-up block, with the GPU synchronized around each measurement. Because a block is emitted in a single call, MOSPA has no per-step time to report, and the per-step columns of Table~\ref{tab:streaming_results} are left blank. The real-time factor is defined for both methods as computation time divided by the duration of motion produced, so that column is directly comparable.

\paragraph{Throughput calculation.}
One 240-frame block spans 8.0\,s of motion and takes 2{,}857\,ms on the RTX A6000 and 11{,}415\,ms on the TITAN RTX, giving real-time factors of 0.357 and 1.427. Our online configuration reaches 0.376 and 0.885 on the same hardware. RTF measures computational throughput, independently of the input waiting time required before generation.

The gap widens with weaker hardware. Moving from the RTX A6000 to the TITAN RTX multiplies MOSPA's block time by 3.99 and our online step time by 2.35. Per second of generated motion, MOSPA evaluates its network 125 times over the full 240-token sequence, whereas our model takes 7.5 autoregressive steps, each extending the KV cache by three tokens and invoking the rectified-flow head for ten Euler steps.

\paragraph{Initial response latency calculation.}
With the initial motion segment supplied, the acoustic token for the first new block becomes available 142.7\,ms after audio onset (Appendix~\ref{sec:implementation}). Adding the measured mean online generation time of 50.2\,ms gives an estimated response latency of 192.9\,ms on the RTX A6000. For MOSPA's 8.0\,s input window, the corresponding estimate is 10{,}857\,ms, including 2{,}857\,ms of generation. Both estimates combine input waiting with mean generation time, using steady-state computation measurements. Acoustic preprocessing and encoding, initialization, rendering, and output delivery are excluded.

%% file: sections/appendix_modular_spatial.tex
\section{Generation without SSL Inputs}
\label{sec:modular_spatial_conditioning}

\paragraph{Motivation and setting.}
Sonic-AMG separates the source of spatial conditions from the autoregressive motion generator.
In our SAM setting, source-location annotations are available alongside binaural audio~\citep{xu2025mospa} and provide the geometry used by the main model.
At deployment, the source position can instead be supplied by an upstream sound source localization module, after transformation into the required coordinate frame, or by scene metadata in a virtual environment.
To examine an alternative realization of this interface, we instantiate a variant that derives spatial tokens directly from binaural audio.
We refer to this variant as \emph{No-SSL}, denoting the absence of source-location inputs at inference.
It retains the initial four-frame motion segment and the reaction-style label used by the main model.

\begin{figure}[htbp]
 \centering
 \includegraphics[width=\linewidth]{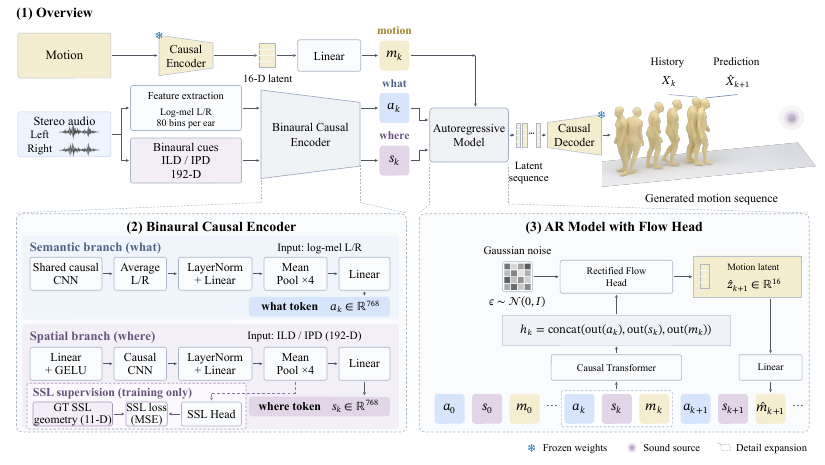}
 \caption{\textbf{Sonic-AMG with audio-derived spatial conditioning.} The semantic and spatial branches derive \emph{what} and \emph{where} tokens from log-mel and ILD/IPD features, respectively. These tokens use the same interface to the autoregressive motion generator. The training-only SSL head predicts the normalized 11-dimensional geometric descriptor from pooled spatial features, supervised by ground-truth geometry through an MSE loss. Inference uses the learned spatial features directly, without the SSL head or spatial feedback. Motion poses are schematic.}
 \label{fig:pipeline_nossl}
\end{figure}

\paragraph{Audio-derived spatial tokens.}
Figure~\ref{fig:pipeline_nossl} shows the variant's architecture.
For each audio frame, we concatenate interaural level differences (ILD) and the sine and cosine of interaural phase differences (IPD) over 64 frequency bands, giving a 192-dimensional input.
A linear projection with GELU, a two-layer causal CNN, and a layer-normalized linear projection encode these cues.
Mean pooling over four frames produces a spatial feature $\mathbf{f}^{s}_k\in\mathbb{R}^{128}$ at the motion-latent rate.
The spatial token is $\mathbf{s}_k=W_s\mathbf{f}^{s}_k+\mathbf{e}_c$, where $W_s$ projects to the transformer width of 768 and $\mathbf{e}_c$ is the reaction-style embedding.
The semantic branch encodes log-mel features with a shared per-ear causal CNN, averages the left and right streams, and applies projection and pooling to obtain the acoustic token.
Unlike the main encoder, this variant omits the additional causal CNN after left/right averaging in the semantic branch.

\paragraph{Motion generation.}
The variant preserves the interleaved sequence $[\mathbf{a}_k,\mathbf{s}_k,\mathbf{m}_k]$, the causal transformer, the rectified-flow head, and the frozen causal motion autoencoder.
We train the generator jointly with the audio encoders using the teacher-forced and 16-step rehearsal objectives described in Section~\ref{sec:rollout}.
Both conditioning streams are computed from the audio, while sampled motion latents extend the autoregressive history.
This variant therefore does not decode root poses to recompute spatial conditions during rehearsal or inference.
Its spatial conditions follow the recorded binaural stream rather than an explicit source position updated relative to the generated body.

\paragraph{Geometric supervision during training.}
No-SSL retains source-location annotations as training targets.
An auxiliary head $H_{\mathrm{SSL}}$ predicts the normalized 11-dimensional descriptor of Eq.~(\ref{eq:spatial_descriptor}) from $\mathbf{f}^{s}_k$, before its projection to a spatial token.
For valid training positions $\mathcal{V}$, the SSL objective is
\begin{equation}
 \mathcal{L}_{\mathrm{SSL}}
 =\frac{1}{11|\mathcal{V}|}\sum_{k\in\mathcal{V}}
 \left\|H_{\mathrm{SSL}}(\mathbf{f}^{s}_k)
 -\frac{\boldsymbol{\phi}^{\mathrm{GT}}_k-\boldsymbol{\mu}_{\phi}}{\boldsymbol{\sigma}_{\phi}}\right\|_2^2,
 \label{eq:nossl_ssl_loss}
\end{equation}
where normalization statistics are computed from the training set.
The semantic features also support audio energy, onset, and activity regression with loss $\mathcal{L}_{\mathrm{event}}$, alongside audio-class classification with loss $\mathcal{L}_{\mathrm{class}}$.
The complete training objective is
\begin{equation}
 \mathcal{L}_{\mathrm{No\text{-}SSL}}
 =\tfrac12(\mathcal{L}_{\mathrm{TF}}+\mathcal{L}_{\mathrm{reh}})
 +0.10\mathcal{L}_{\mathrm{SSL}}
 +0.05\mathcal{L}_{\mathrm{event}}
 +0.05\mathcal{L}_{\mathrm{class}}.
 \label{eq:nossl_total_loss}
\end{equation}
The auxiliary heads are used only for training.
The predicted geometric descriptor is not passed to the motion generator: at inference, the learned spatial feature directly supplies the \emph{where} token.

\paragraph{Scope of the experiment.}
This configuration instantiates the same motion-generation interface with an audio-derived provider of spatial conditions.
It removes the need for source-position inputs at inference while retaining geometric supervision during training.

\paragraph{Quantitative comparison.}
No-SSL must infer spatial cues from incoming binaural audio while generating motion causally, without source-location inputs or access to future audio.
Despite these demands, it surpasses EDGE in motion fidelity and achieves conditioning alignment and diversity comparable to MOSPA (Table~\ref{tab:nossl_results}).
These results show that audio-derived spatial tokens provide effective conditioning for autoregressive reactions.
Sonic-AMG can therefore support streaming motion generation in scenarios where explicit source geometry is unavailable.

\begin{table}[htbp]
 \centering
 \caption{\textbf{No-SSL comparison on SAM.} Baselines are reused from Table~\ref{tab:reference_results}. $\dagger$ denotes results from \citet{xu2025mospa}.}
 \label{tab:nossl_results}
 \small
 \input{tables/nossl_results}
\end{table}

%% file: tables/nossl_results.tex
\begin{tabular*}{\linewidth}{@{\extracolsep{\fill}}lrrrrr@{}}
\toprule
Method & FID$\downarrow$ & R@1$\uparrow$ & R@2$\uparrow$ & R@3$\uparrow$ & Diversity$\rightarrow$ \\
\midrule
GT motion$^{\dagger}$ & 0.001 & 1.000 & 1.000 & 1.000 & 23.616 \\
\midrule
EDGE$^{\dagger}$ & 13.993 & 0.886 & 0.960 & 0.977 & 23.099 \\
POPDG$^{\dagger}$ & 20.967 & 0.762 & 0.886 & 0.934 & 22.536 \\
LODGE$^{\dagger}$ & 102.289 & 0.444 & 0.594 & 0.679 & 21.101 \\
Bailando$^{\dagger}$ & 168.396 & 0.077 & 0.134 & 0.182 & 17.347 \\
MOSPA & \textbf{8.242} & \textbf{0.942} & \textbf{0.984} & \textbf{0.994} & \textbf{23.412} \\
\midrule
Sonic-AMG (No-SSL) & \underline{11.917} & \underline{0.915} & \underline{0.973} & \underline{0.987} & \underline{23.371} \\
\bottomrule
\end{tabular*}